\documentclass[journal,twoside,web]{ieeecolor}
\usepackage{generic}
\usepackage{cite}
\usepackage{amsmath,amssymb,amsfonts}
\usepackage{algorithmic}
\usepackage{graphicx}
\usepackage{textcomp}
\usepackage[utf8]{inputenc}
\usepackage{graphicx,subcaption}
\usepackage{amssymb}
\usepackage{amsmath}
\usepackage{makecell}
\usepackage[ruled,linesnumbered]{algorithm2e}
\usepackage{longtable}

\usepackage{array,multirow,graphicx}

\usepackage{amsmath}
\DeclareMathOperator*{\argmax}{argmax}
\DeclareMathOperator*{\argmin}{argmin}

\graphicspath{ {./Figures/} }
\usepackage{widetext}

\usepackage{float}

\DontPrintSemicolon
\newcommand{\nosemic}{\renewcommand{\@endalgocfline}{\relax}}
\newcommand{\dosemic}{\renewcommand{\@endalgocfline}{\algocf@endline}}
\let\oldnl\nl
\newcommand{\nonl}{\renewcommand{\nl}{\let\nl\oldnl}}

\usepackage{mathtools}

\usepackage{xcolor}
\usepackage{soul}

\newenvironment{flushenum}{
\begin{enumerate}
  \setlength{\leftmargin}{-4pt}
}{\end{enumerate}}

\DeclarePairedDelimiterX{\infdivx}[2]{(}{)}{%
  #1\;\delimsize\|\;#2%
}

\DeclarePairedDelimiter{\norm}{\lVert}{\rVert}

\DeclareMathAlphabet{\mathcalligra}{T1}{calligra}{m}{n}

\def\BibTeX{{\rm B\kern-.05em{\sc i\kern-.025em b}\kern-.08em
    T\kern-.1667em\lower.7ex\hbox{E}\kern-.125emX}}
\begin{document}
\title{A Knowledge Distillation Ensemble Framework for Predicting Short and Long-term Hospitalisation Outcomes from Electronic Health Records Data}

\author{Zina M Ibrahim, Daniel Bean, Thomas Searle, Linglong Qian, Honghan Wu, Anthony Shek, Zeljko Kraljevic, James Galloway, Sam Norton, James T Teo,  Richard JB Dobson
\thanks{Submitted on 18/11/2020; revised on 26/03/2021 and 22/05/2021; accepted on 06/06/2021. ZI and RJBD are supported by (1) NIHR Biomedical Research Centre at SLaM and King's College London, London, U.K. and (2) NIHR University College London Hospitals Biomedical Research Centre. RJBD is further supported by (1) Health Data Research (HDR) UK and (2) The BigData@Heart Consortium under grant agreement No. 116074. AS is supported by a King's Medical Research Trust studentship. DMB is funded by a UKRI Innovation Fellowship as part of Health Data Research UK MR/S00310X/1. HW is supported by MRC and HDR UK Grant (MR/S004149/1) and Wellcome Institutional Translation Partnership Award (PIII054). JTHT is supported by London AI Medical Imaging Centre for Value-Based Healthcare (AI4VBH) and NIHR Applied Research Collaboration South London at King's College Hospital NHS Foundation Trust. }
\thanks{ZI, DB, TS, LQ, ZK  and RJBD are with the Department of Biostatistics and Health Informatics, King's College London, SE5 8AF UK (\{zina.ibrahim; daniel.bean; thomas.searle; linglong.qian; zeljko.kraljevic; richard.j.dobson\}@kcl.ac.uk). HW, RJBD and ZI are with the Institute of Health Inforamtics, University College London, UK ( \{honghan.wu; r.dobson; z.ibrahim\}@ucl.ac.uk). AS is with the Department of Clinical Neuroscience, King's College London, London, UK (anthony.shekh@kcl.ac.uk). JG is with the Centre for Rheumatic Diseases, King's College London (james.galloway@kcl.ac.uk). SN is jointly appointed by the Department of Psychology and the Department of Inflammation Biology, King's College London ( sam.norton@kcl.ac.uk). JT is with King's College Hospital NHS Foundation Trust (jamesteo@nhs.net).}}

\maketitle

\begin{abstract}
The ability to perform accurate prognosis is crucial for proactive clinical decision making, informed resource management and personalised care. Existing outcome prediction models suffer from a low recall of infrequent positive outcomes. We present a highly-scalable and robust machine learning framework to automatically predict adversity represented by mortality and ICU admission and readmission from time-series of vital signs and laboratory results obtained within the first 24 hours of hospital admission. The stacked ensemble platform comprises two components: a) an unsupervised LSTM Autoencoder that learns an optimal representation of the time-series, using it to differentiate the less frequent patterns which conclude with an adverse event from the majority patterns that do not, and b) a gradient boosting model, which relies on the constructed representation to refine prediction by incorporating static features. The model is used to assess a patient's risk of adversity and provides visual justifications of its prediction. Results of three case studies show that the model outperforms existing platforms in ICU and general ward settings, achieving average Precision-Recall Areas Under the Curve (PR-AUCs) of 0.891 (95$\%$ CI: 0.878 - 0.939) for mortality and 0.908 (95$\%$ CI: 0.870-0.935) in predicting ICU admission and readmission. 

\end{abstract}



\section{Introduction}

The secondary re-use of routinely collected patient data has been a facilitator of innovations aiming to improve patient care. A prominent example is the development of early warning systems that predict adversity from patient physiological measurements. The majority of early warning models take the form of ad-hoc scoring tools \cite{earlywarning} such as the National Early Warning Score (NEWS2) widely used in the United Kingdom \cite{news2}. Such tools estimate a patient's risk of adversity using aggregates of physiological measurements \cite{news2performancelalala}. Generally, scoring tools suffer from low sensitivity due to overlooking the dependencies among the temporal signatures underlying a patient's physiology. Machine Learning models have been developed to overcome the limitations of scoring tools via sophisticated architectures that capture non-linearities within the multivariate temporal patient data \cite{ml3}.

Despite the promising results of Machine Learning early warning systems, we find that existing approaches bear several shortcomings that adversely affect model performance and adoption potential. First, generic clinical outcome prediction models are scarce. Most existing models are condition-specific, being developed and evaluated with a single condition in mind, e.g. sepsis \cite{sepsis}, cardiac patients \cite{cardio2}, COVID-19 \cite{covidmodel}, brain injury \cite{brain}. The few condition-agnostic models available are mostly limited to intensive care (ICU) settings where the magnitude of measurements is high, and the population is more uniform in acuity levels \cite{icu2,icu3}. Second, predicting adverse clinical outcomes is an imbalanced learning problem because any adverse outcome is only present in a minority of the patient sample used to train and evaluate a model. To illustrate, consider the United Kingdom's in-hospital mortality rates, which are around  $23\%$ in ICU settings \cite{mortalityicu} and  $4\%$ in secondary wards \cite{mortalityward}. Similarly, cardiac arrest incidence is estimated to be $2.3\%$ of ICU admissions \cite{cardiacincidence}. Nevertheless, most current adversity-prediction models are benchmarked using the Area Under the Curve (ROC-AUC) \cite{attention3shamout, mortalityprediction1, mortalityprediction2,mortalitypredictionlatest, icu1}, which is known to overestimate model performance on minority outcomes under imbalanced distributions \cite{imbalanced2,cstat}. The result is a general over-optimism in existing models’ performance. Finally, in contrast to medical practice, where both the dynamics of a patient's physiology and personal characteristics (e.g. demographics) are used for prognosis, clinical outcome predictions models seldom consider the interplay between the dynamic and static data available about a patient. Existing models either consider the two views descriptively and not a predictive context \cite{brajer}, or distinctly without consideration of the interplay between the two \cite{shamoutcovid}. 

\begin{figure}[hbt!]
 \centering
\includegraphics[width=.5\textwidth,keepaspectratio]{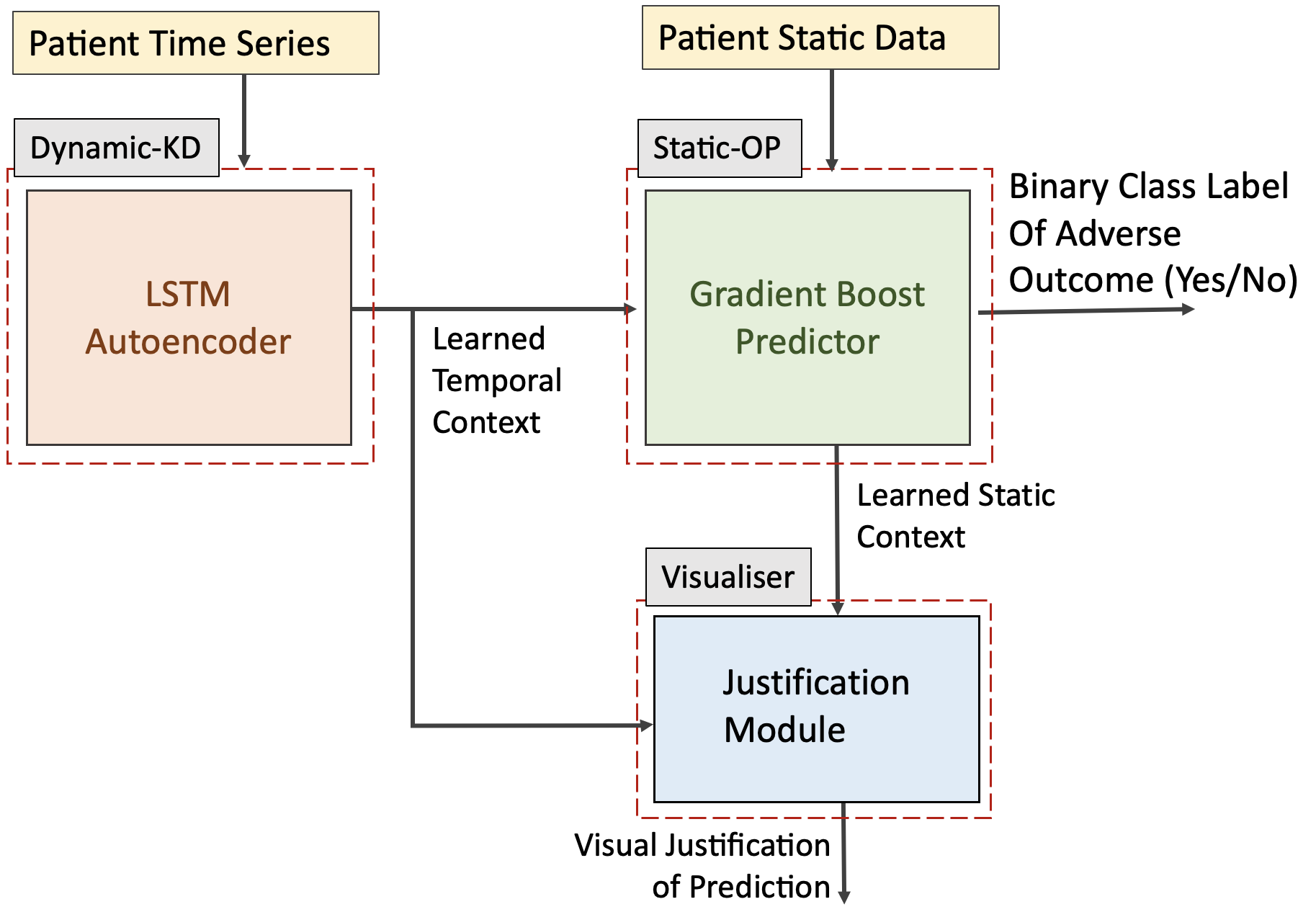}\caption{\label{fig:architecture} The Knowledge-Distillation Outcome Predictor \emph{(KD-OP)}. The Dynamic Knowledge Distiller \emph{(Dynamic-KD)} module learns a compressed temporal representation of the no-adversity class. Static Outcome Predictor \emph{(Static-OP)} performs the final prediction by combining \emph{Dynamic-KD}'s output with patient static features. The framework's visualisation module uses both contexts to justify the predictions made. }
\end{figure}

This paper presents \emph{KD-OP} (Knowledge Distillation Outcome Predictor), an ensemble Machine Learning framework designed to overcome the current difficulties in predicting adverse clinical outcomes from electronic health records data. The framework (Figure \ref{fig:architecture}) comprises two learner modules. The first, \emph{Dynamic-KD} (Dynamic Knowledge Distiller), learns from the multivariate time-series of a patient's physiology, while the second, \emph{Static-OP} (Static Outcome Predictor), estimates the risk of adversity using static features (e.g. demographics and aggregate measurements). \emph{KD-OP} uses a \emph{stacked} architecture to capture the interplay between the two patient views, using \emph{Dynamic-KD}'s learned context to guide the predictions made by \emph{Static-OP}. In contrast to existing clinical outcome prediction models, \emph{KD-OP} is designed to befit the relative infrequency of adverse outcomes in real hospital data. This is achieved by implementing \emph{Dynamic-KD} as a Long Short-term Memory (LSTM) Autoencoder, thereby reformulating the prediction task into one of \textit{outlier detection}, whereby adverse outcomes (e.g. mortality =1) are modelled as outliers. We evaluate the framework's predictive power under ICU and general ward settings, using metrics specifically designed for imbalanced classification models \cite{rocbad,pr}. The stacked architecture generates visual justifications of its predictions based on the learned temporal and static context.

 \section{Related Work}

Ensemble models have shown superior performance compared to single-classifier architectures in predicting hospitalisation outcomes \cite{icumortality,attention3shamout,mortalityprediction2,mortalitypredictionlatest,brazilian,super}. The advantages of ensemble models have been realised either by capturing different data modalities (i.e. static and dynamic views) \cite{shamoutcovid}, or by consolidating predictions by several 'weak' classifiers \cite{ensembleread}. However, in all existing clinical outcome prediction frameworks, the ensemble's learners are linked by additively combining their respective predictions, e.g. via weighted averaging. This approach is problematic for ensembles operating on different data modalities because the final prediction is not depictive of the possible dependencies between the data's static and the dynamic views \cite{combiningstaticdynamic}. There is, therefore, a missed opportunity in developing models that capture the interplay between a patient's dynamically changing physiology and personal characteristics.
 Moreover, additive ensembles have been shown to fail to alleviate the individual learners' bias; they are generally outperformed by alternative models that \emph{stack} strong classifiers into an ensemble \cite{votingvsstacking,combiningstaticdynamic}.

The proposed stacked architecture dedicates its time-series prediction task to an LSTM Autoencoder. The model choice is motivated by the natural marginal representation of adverse outcomes in hospital data and the known ability of LSTM AutoEncoders to identify minority outcomes from imbalanced time-series \cite{lstmoutlier,lstmoutlier2}. The idea is that an LSTM Autoencoder encodes the time-series in a low dimensional representation capturing its most informative features. The compact representation then only enables the reconstruction of the representative features from new input without noise. By strictly training the LSTM Autoencoder on the majority (no-adversity) time-series, outliers (time-series corresponding to adverse outcomes) will generate high reconstruction errors \cite{lstmoutlier2}. LSTM Autoencoders have been effectively used in fall detection \cite{fall}, sensor failure prediction \cite{sensor}, fraud detection \cite{fraud} and video surveillance \cite{surveillance}. LSTM Autoencoders have also shown great potential in healthcare \cite{application}, with applications in retinal eye research \cite{retinal}, patient subtyping \cite{subtyping} and healthcare fraud detection \cite{fraudhealthcare}.

\section{Methodology}

In this work, the prediction of adverse clinical outcomes is expressed as a binary classification whose input comprises two types of multivariate data collected for a patient: 1) dynamic data, comprising physiological measurements and laboratory results routinely collected at the bedside either automatically or by healthcare practitioners (e.g. heart rate), and 2) static data, which is either recorded during admission (e.g. demography), summarised during examination (e.g. comorbidities) or is a dynamic-data aggregate habitually used by healthcare practitioners to evaluate a patient's state (e.g. maximum heart rate during in a given day). For a given adverse outcome, the framework (Figure \ref{fig:architecture}) assigns each patient a binary class label using the predicted probability of the patient's risk and an adversity threshold $\gamma \in [0,1]$, which the model also learns. 

The dynamic data, which takes the form of a multivariate time-series, is processed by the Dynamic Knowledge Distiller \emph{(Dynamic-KD)} module to learn a vector representation assigning each patient a risk score based on the patient's temporal signatures alone. \emph{Dynamic-KD}'s learned risk is used to guide the prediction process of the Static Outcome Predictor \emph{(Static-OP)} module, a Gradient Boost classifier that uses the static view to predict the final class label. The ensemble model is designed to improve predictive power and adoption potential in the following ways:

\begin{flushenum}
\item By adopting a \emph{stacked} architecture to capture the interplay between the temporal signatures of a patient's physiology and static features, thereby enabling the temporal risk learned by \emph{Dynamic-KD} to drive classification based on static using \emph{Static-OP}.

\item By incorporating a robust mechanism to address the marginal representation of adverse outcomes (e.g. ICU admission = 1) compared to typical outcomes (e.g. no ICU admission) in the overall population. This is achieved by designing the dynamic-learner module, \emph{Dynamic-KD}, using the successful LSTM-Autoencoder outlier detection architecture.
\end{flushenum}

\subsection{Glossary of Terminology and Model Formulation }
For a patient $k$ having $v$ dynamic features each measured over $T$ consecutive time windows, and $u$ single-measurement static features, we represent the patient's dynamic view by a matrix $\boldsymbol{X_k} \in  {\Bbb R}^{T\times v}$, containing the totality of the patient's dynamic observations. Moreover, the patient's static view is represented by a vector $\boldsymbol{x_k} \in {\Bbb R}^u$ consisting of a sequence of $u$ static and aggregate features. For a population of $n$ patients, the ensemble model therefore accepts two types of inputs: the combined multivariate time-series of the $n$ patients, $\boldsymbol{D^d} = \displaystyle{\{{\boldsymbol{X}_k\}}_{k=1}^n}$, and a static patient-feature matrix $\boldsymbol{D^s} = \displaystyle{\{\boldsymbol{x}_k\}_{k=1}^n}$. Additionally, because \emph{Static-OP} is a classifier, it is trained using  $\boldsymbol{D^s}$ and a vector representing the true incidence of the adverse outcome in the cohort $\boldsymbol{y} = \displaystyle{\{y_k\}_{k=1}^n}$.





  

For a given adverse outcome, the goal of the framework is to predict the vectors $\boldsymbol{\hat{p}}  =  \displaystyle{\{\hat{p}_k\}_{k=1}^n}$ and  $\boldsymbol{\hat{y}}  =  \displaystyle{\{\hat{y}_k\}_{k=1}^n}$. Each $\hat{y}_{k}   \in \{0,1\}$ is a binary variable representing the predicted onset of the adverse outcome for a single patient; ${\hat{p}}_k$ is the predicted probability of the outcome for one patient, which we retain for use during the interpretation stage. Naturally, the class distribution of $\boldsymbol{y}$ is highly imbalanced in favour of non-adverse outcome, as will be demonstrated in the evaluation sections. The framework learns the probability of an adverse outcome for the population of $n$ patients, $ \mathcalligra{\boldsymbol{\hat{p}}}_k$ from the two views of the clinical data $\boldsymbol{D}^d$ and $\boldsymbol{D}^s$, using it to estimate $\boldsymbol{\hat{y}_k}$. 
   
   \subsection{Data Processing} \label{sec:processing}
   
Data extracted from Electronic Health Records (EHRs) is generated as a by-product of routine clinical care. As a result, the variables making up a patient's dynamic data view comprise irregularly-sampled time sequences of physiological measurements. Using the extracted data to train \emph{KD-OP} entailed transforming the irregular time series into $\boldsymbol{D^d = \{ X_1, ...X_n\} }$ for the $n$ patients, where each $\boldsymbol{X_k}$ is a $T \times v$ matrix, of fixed number $T$ of equally-sized observation windows and $v$ vital signs measured at each time window $t \in T$. Furthermore, in order for $\boldsymbol{D^d}$ to be digestible by the two Machine Learning models, interpolation of missing data and scaling are required.

To alleviate missingness while overcoming non-uniform sampling, the number of observation windows $T$ used to construct $\boldsymbol{D^d}$ was optimised iteratively by maximising completeness while minimising the length of the observation windows $t \in T$. If multiple observations are present for a variable during a window, they are binned using a knowledge-based approach that mimics the summaries used by clinicians in practical settings (Table \ref{tab:aggregation}). 

\begin{center}
\begin{table}[ht!]
\resizebox{\linewidth}{!}{%
\begin{tabular}{ l l } 
 \hline
    \textbf{Aggregation Function} & \textbf{Variables} \\\hline
	Minimum &Albumin, Central Venous Pressure, Haemoglobin \\\hline
	Maximum &Alanine Aminotransferase (ALT), Assured shorthold \\
					&tenancies( AST), Bilirubin, Creatinine , Fibrinogen\\
					& C-Reactive Protein , Lactate Dehydrogenase\\
					& International Normalized Ratio (INR)\\
					&Partial Pressure of Oxygen Dioxide (PaCO$_2$), \\
					&Partial Pressure of Oxygen PaO$_2$\\
					&Partial Venous Pressure of Oxygen Dioxide PvCO2\\
					&Peripheral O$_2$ Saturation, oxygen saturation (SaO$_2$)\\
					&Spontaneous Respiratory Rate, Urea  \\\hline

	Average & Lymphocytes , Neutrophils, NEWS2, Platelets  \\\hline

	Minimum, Maximum &Arterial Oxygen Content (CaO$_2$)\\
								 &Centralvenous O$_2$ Saturation, Diastolic Blood Pressure  \\
								 &Heart rate,  Inspired Fraction of Oxygen (FiO$_2$) \\
								 &Mean Blood Pressure, PO$_2$/FiO$_2$,  Systolic Blood Pressure \\
								 & Temperature  , White Blood Cell Count  \\\hline								 
\end{tabular}}
\caption{\label{tab:aggregation} Variable binning functions used in this work. When a variable is associated with more than one binning function (e.g. heart rate), then two features are generated}.
\end{table}
\end{center}

Similarly to \cite{attention3shamout}, we imputed the resulting time-series Gaussian Process Regression (GPR).  Gaussian Processes (GPs) extend the assumptions of a Gaussian distribution over functions, whereby a function $f$ is a Gaussian Process if it is entirely characterized by its mean $m$ and covariance $o$ functions: $f \thicksim \mathcal{GP}(m, o)$ \cite{gaussian}. GPR thus provides a non-parametric method for accommodating statistical uncertainty measures in a regression problem. It has been used in a wide range of applications and has been found to produce superior results when used to interpolate missing data in multivariate time-series \cite{clifton,gaussianImpute}.

We defined a GPR over the dynamic features in $\boldsymbol{D^d}$ for the $T$ time windows. To ensure the interpolated values are meaningful, we grouped patients into subpopulations based on demographical similarities (age, sex, co-morbidity indices), subsequently defining the mean function of each feature as the subpopulation mean $\mu$ of patients within the same demographical group. We used the exponential covariance function, resulting in the following definition of our GPR:
\begin{equation*}
f \thicksim \mathcal{GP}(m, o): m(x) = \mu \,\,\, and  \,\,\,  o(x, x') = \exp(\displaystyle{-\frac{1}{2}(x-x')^2)} 
\end{equation*}

Furthermore, the data was normalised using a number of scaling techniques including absolute-value scaling and min-max scaling. Using a min-max scaler with a range of $[0,1]$ achieved the best classification performance and was therefore adopted for all the models. The interpolation and normalisation models were both fitted to the training data, using each model's learned parameters to process the test data. 


    \subsection{The Architecture}
    
\begin{figure}[ht]

\centering
\includegraphics[width=.45\textwidth]{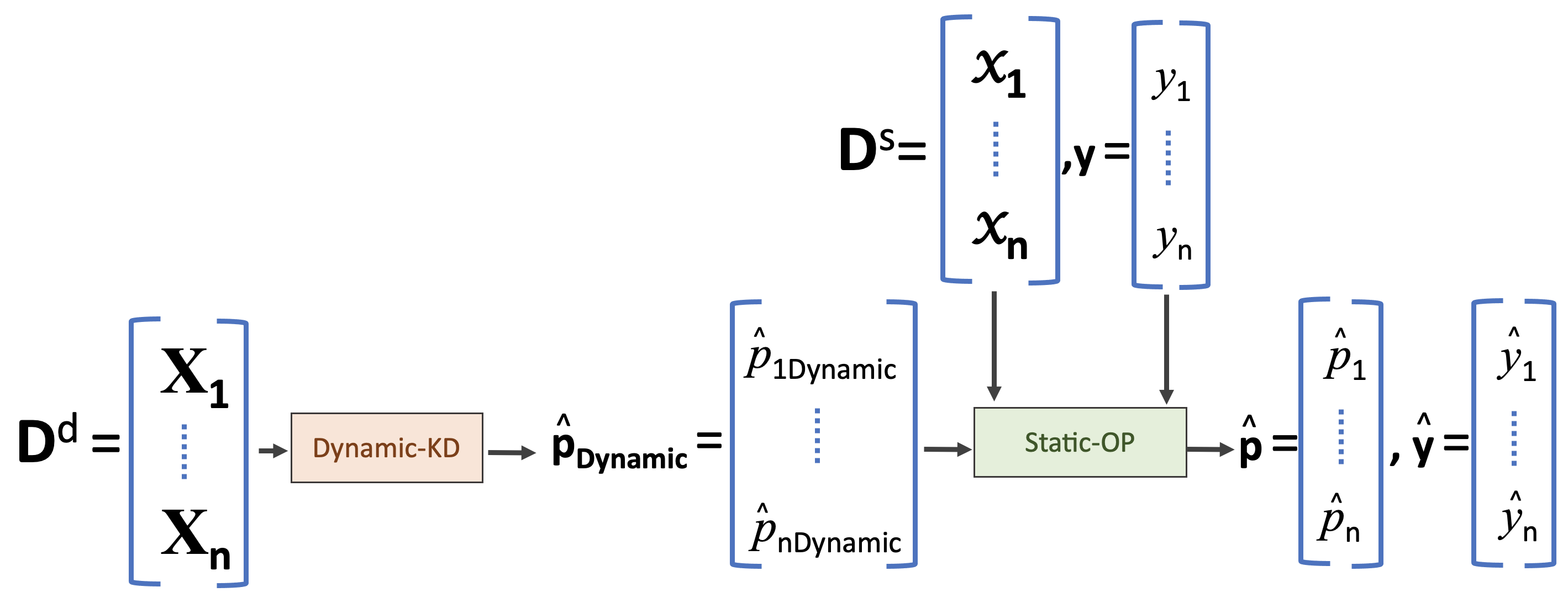}

\caption{\emph{KD-OP} Data Flow: $\boldsymbol{D^d}$ contains the totality of the dynamic data for $n$ patients, where each $\boldsymbol{X_k}$ is a $T\times v$ matrix corresponding to the time series of a single patient with $v$ dynamic features over $T$ time windows. $\boldsymbol{\hat{p}_{Dynamic}} $ is a vector of length $n$ containing, for each patient $k$, the likelihood of the patient $k$ of belonging to the adversity class based on the learned temporal context embedded within $\boldsymbol{X_k}$. \emph{Static-OP} uses $\boldsymbol{\hat{p}_{Dynamic}} $ to guide classification based on data containing static features, $\boldsymbol{D^s}$, to output the final predicted probabilities $\boldsymbol{\hat{p}} $ and adversity classes $\boldsymbol{\hat{y}} $. }
\label{fig:terminology}

\end{figure}

Figure \ref{fig:terminology} illustrates the data flow of the proposed architecture. As the figure shows, the combined multivariate time-series of the $n$ patients, $\boldsymbol{D^d}$ is used as input to \emph{Dynamic-KD}, which is an unsupervised LSTM Autoencoder that learns a vector  ${\boldsymbol{\hat{p}_{Dynamic}}} \in [0,1]$ from the time-series such that for two individuals $a$ and $b$ whose time-series observations $\boldsymbol{X_a}$ and $\boldsymbol{X_b}$ $\in$ $\boldsymbol{D^d}$, if  $\hat{y}_{a} > \hat{y}_{b}$ then:
\begin{equation*}
   \log (\hat{p}_{a Dynamic} ) >  \log(\hat{p}_{b Dynamic} ) 
\end{equation*}

That is, the vector representation $\boldsymbol{\hat{p}}_{Dynamic}$ creates a separation between time-series corresponding to adverse outcomes and those corresponding to normal outcomes. The  use of log transformation captures the difference in the order of magnitude of the resulting representations, rather than possibly non-significant fluctuations within the actual values \cite{logtransform}.

The second module, \emph{Static-OP}  is a classification ensemble based on gradient boost trees. \emph{Static-OP} is trained using $\boldsymbol{D^s}$ and the true incidence of the outcome in the cohort, $\boldsymbol{y}$ to estimate the final prediction probability $\boldsymbol{\hat{p}}$. During the training process of \emph{Static-OP}, $\boldsymbol{\hat{p}_{{Dynamic}}}$ is used as sample weights to guide training. The two modules form the bi-level stacked classification system \emph{KD-OP} (Knowledge Distillation Outcome Predictor). The output for the ensemble is: 
\begin{equation}
\boldsymbol{\hat{y}} =  \begin{cases}
0 &\text{if } \boldsymbol{\hat{p}_{Static}} > \gamma\ \\
1 &  \text{otherwise}
\end{cases}
\label{eq:yhat}\end{equation}
Where the adversity threshold $\gamma \in [0,1]$ is a learned parameter corresponding to the optimal threshold for classification selected by optimising the mean Area Under the Precision-Recall Curve (PR-AUC). The remainder of this section details the design of \emph{KD-OP}'s individual modules.

\subsubsection{Dynamic-KD}

The first module is designed to use the data's dynamic view, i.e. the multivariate time-series input $\boldsymbol{D^d}$, to learn a vector-form separation between the majority (non-adverse outcome) time-series and time-series corresponding to adverse events. The module capitalises on the significantly lower frequency of adverse outcomes (e.g. ICU admission = 1) compared to typical outcomes (e.g. no ICU admission) in the overall population, enabling viewing adverse outcomes as outliers.

\begin{figure}[hbt!]
 \centering
 \includegraphics[width=.5\textwidth,keepaspectratio]{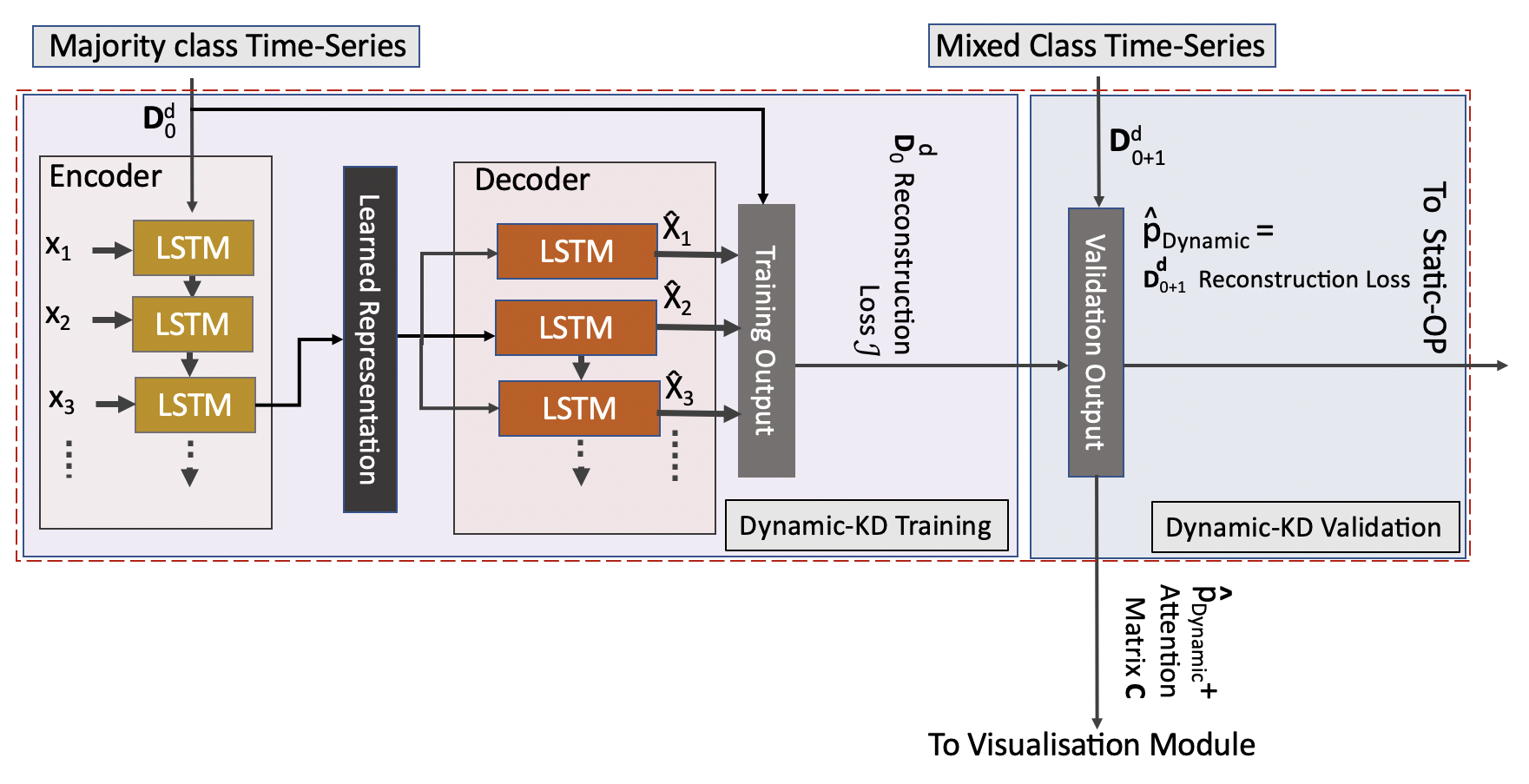}\caption{\label{fig:DynamicKD} The Dynamic-KD LSTM Autoencoder Module. $\boldsymbol{D^d_0}$ refers to the portion of the time-series data which corresponds to patients with typical (non-adverse) outcomes used to self-train the LSTM Autoencoder;   $\boldsymbol{D^d_{0+1}}$ refers to mixed-class time-series. When provided with $\boldsymbol{D^d_{0+1}}$ as input, the trained LSTM Autoencoder, yields $\boldsymbol{\hat{p}_{Dynamic}}$, which is representative of the probability of each patient $k$ in $\boldsymbol{D^d_{0+1}}$ belonging to the minority class.}
\end{figure}

\emph{Dynamic-KD} (Figure \ref{fig:DynamicKD}) consists of an unsupervised LSTM-Autoencoder architecture which is 'self-trained' using a subset of the time-series containing only majority outcomes, $\boldsymbol{D^d_0}$. For each patient matrix $\boldsymbol{X}_k$ in $\boldsymbol{D^d_0}$, the training procedure aims to reconstruct $\boldsymbol{X}_k$'s input sequence by minimising a distance-based objective function $\mathcal{J}$. $\mathcal{J}$ is the reconstruction loss measuring the difference between the input vectors in the original series $\boldsymbol{X}_k$ and the vectors of the reconstructed series $\boldsymbol{\hat{X}}_k$. $\mathcal{J}$ is defined as below: 
\begin{equation}
		\mathcal{J} = \displaystyle\sqrt{ \sum_{t = 1}^T  \norm{\boldsymbol{x_t} -\boldsymbol{ \hat{x_t}} }_2^2}
\label{eq:loss} \end{equation}

Where $T$ is the number of multivariate observations for each patient $k$, $\boldsymbol{x_t}$ and $\boldsymbol{\hat{x}_t}$ are the multi-feature vectors at time $t$,  and $\norm{.}_2$ is the L2-norm. Once trained, the LSTM Autoencoder is run on a subset of the mixed-class time series ($\boldsymbol{D^d_{0+1}}$ in the figure). Because training was performed strictly on majority-class series, the values of the reconstruction loss $\mathcal{J}$ associated with outliers (samples corresponding to the adversity class) in $\boldsymbol{D^d_{0+1}}$ will be significantly higher than the reconstruction loss $\mathcal{J}$ associated with the majority (no adversity) samples of $\boldsymbol{D^d_{0+1}}$. As a result, the vector output $\boldsymbol{\hat{p}_{Dynamic}}$, which is the reconstruction loss associated with $\boldsymbol{D^d_{0+1}}$, will be representative of the probability of each patient $k$ in $\boldsymbol{D^d_{0+1}}$ belonging to the minority (outlier) class.

\begin{figure}[hbt!]
 \centering
 \includegraphics[width=.8\columnwidth,keepaspectratio]{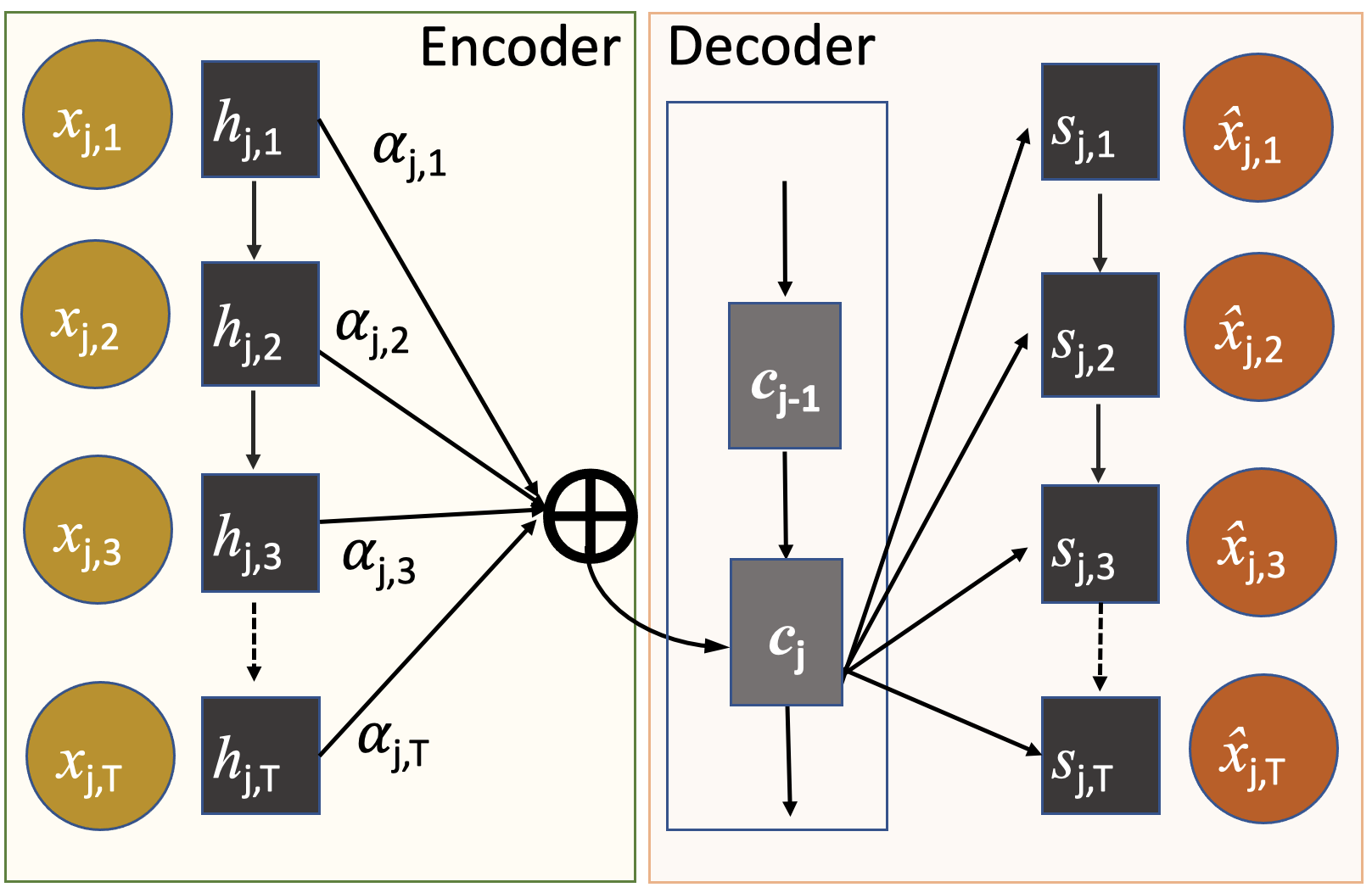}\caption{\label{fig:encoderdecoder} The attention mechanism of the encoder-Decoder Architecture demonstrated at feature level. $X$: input sequence; $\hat{X}$: output sequence; $h$: encoder hidden state; $s$: decoder hidden state; $c$ attention (context) vector. }
\end{figure}

 \emph{Dynamic-KD} adopts an attention mechanism over the time steps to capture the most important features in each sequence as proposed by \cite{attention} and successfully implemented in \cite{attention1, attention2,attention3shamout}. Figure \ref{fig:encoderdecoder} shows a feature-level representation of the attention mechanism in the encoder-decoder architecture of \emph{Dynamic-KD},  reconstructing a multi-variable sequences over $T$ time step batches (i.e. $T$ ordered sequences per patient). For each feature $j$, a soft attention mechanism is implemented over the encoder's hidden states to obtain a distinct context vector $\boldsymbol{c_j}$. $\boldsymbol{c_j}$ attenuates the most informative hidden states in $s_{j,1}, .... s_{j,T}$ of the decoder based on the encoder's latent representation and is computed as follows:

For each feature $j$ , the attention probabilities based on the encoded sequence $\boldsymbol{\alpha} = (\alpha_1, ...., \alpha_T)$  are calculated using the encoded sequence and the encoder's internal hidden states. First, the importance of information at each time step for feature $j$ is calculated:  

\begin{equation*}
\boldsymbol{e_{j,t} = a(\mathcal{U}_j\circledast  \ s_{t-1} +  \mathcal{W}_j \circledast  h_j + b_j)}
\end{equation*}

Where $\boldsymbol{\mathcal{U}_j}$ and $\boldsymbol{\mathcal{W_j}}$ are trainable parameters capturing the input-to-hidden and hidden-to-hidden transitions for a given layer $j$ respectively. Terms $\boldsymbol{\mathcal{W}}_j \circledast \boldsymbol{ h_{t-1}}$ and $ \boldsymbol{\mathcal{U}}_j \circledast \boldsymbol{x_t}$ respectively capture the update from the hidden states at the previous step and the new input. $a$ is the activation function. In the decoder layers, we can measure the importance of the information at each time step for each feature $j$ denoted by $e_{j,t}$ using proximity to $\boldsymbol{\mathcal{U}}_j$. Then $\alpha_{j,t}$ is obtained by normalising $e_{j,t}$ using the softmax operation: 

\begin{equation*}
\alpha_{j,t} = 	\displaystyle \frac{exp(e_j)}{\sum_{t=1}^{T} exp(e_t)}
\end{equation*}

Finally, the context vector for each feature $\boldsymbol{c_j}$ is calculated using the weighted sum of the encoded sequence with the attention probabilities. Intuitively, this vector summarizes the importance of the encoded features in predicting $t^{th}$ sequence:

\begin{equation}
\boldsymbol { c_j }= \displaystyle \sum_{t=1}^{T} \alpha_{j,t} h_{j,t}
\end{equation}

As algorithm \ref{alg:dynamickd} shows, the \emph{Dynamic-KD} LSTM-Autoencoder is trained in batches of $\boldsymbol{D^d_{0}}$ to minimise the reconstruction loss $\mathcal{J}$ (line 1). The resulting loss is therefore representative of the training errors associated with the negative (majority) class.

\begin{algorithm}[ht]
\SetAlgoLined
\SetKwInOut{Input}{Receives} 
\SetKwInOut{Output}{Returns}
\Input{Training and validation subsets of the multivariate, regularly-sampled and batched time-series, $\boldsymbol{D^d}$:  $\displaystyle{\boldsymbol{D}^d_0}$ containing majority-class time-series for training and $ \displaystyle{\boldsymbol{D}^d_{0+1}}$ containing mixed-class time-series for validation}
\Output{Validation reconstruction loss $\boldsymbol{\hat{p}_{Dynamic}}$,   and attention matrix $\boldsymbol{C_{0+1}} $}

\nonl \;
 
  \nonl \textbf{Train Auto Encoder}


	$\displaystyle\hat{\theta}$= $\displaystyle \argmin_{\theta} \,\, \mathcal{J}(\boldsymbol{D^d_{0}})$
	

\nonl \;
 
  \nonl \textbf{Run Auto Encoder}
  
  	 $\boldsymbol{\hat{D}^d_{0+1}}$, $\boldsymbol{C_{0+1}} $ = $Decoder(Encoder(\boldsymbol{D^d_{0+1}}), \hat{\theta})$

	$\boldsymbol{\hat{p}_{Dynamic}}$  =  $\boldsymbol{\hat{D}^d_{0+1}}$ $-$ $\boldsymbol{D^d_{0+1}}$
  	  
 \caption{Dynamic-KD}\label{alg:dynamickd}
\end{algorithm}

Once trained, the LSTM-Autoencoder is run on a non-overlapping subset of the time-series $\boldsymbol{D_{0+1}^d}$ (line 2), which contains mixed data (positive and negative outcomes), using the optimal loss obtained during training. Running the autoencoder yields a reconstruction $\boldsymbol{\hat{D}^d_{0+1}}$ of $\boldsymbol{D^d_{0+1}}$ and an attention matrix $\boldsymbol{C_{0+1}}$. At the end of the procedure, the reconstruction loss $\boldsymbol{\hat{p}_{Dynamic}}$ (line 3) of $\boldsymbol{\hat{D}^d_{0+1}}$ from the original series $\boldsymbol{D^d_{0+1}}$ augments the original highly-dimensional feature space into a linear representation which is descriptive of the deviation from normality (no adversity) with respect to the temporal interactions embedded within the data. $\boldsymbol{\hat{p}_{Dynamic}}$ is, therefore, discriminatory between the two classes and corresponds to the likelihoods of each batch (patient) $k$ belonging to the positive class. $\boldsymbol{\hat{p}_{Dynamic}}$ is used to complement the learning from static features performed by \emph{Static-OP}, while $\boldsymbol{C_{0+1}}$ is fed into the interpretation component of the framework. 

\subsection{Static-OP}
The goal of this module is to complement the predictions made from dynamic data by \emph{Dynamic-KD}. \emph{Static-OP} takes as input static information routinely employed by healthcare practitioners to assess a patient's risk factors (e.g. demographics, symptoms, summary statistics of physiology). In other words, instead of using $\boldsymbol{\hat{p}_{Dynamic}}$ as a predictor of the outcome, it is instead used to drive further classification. This way, the overall pipeline has the advantage of capturing the interplay between dynamic physiological measurements and static features in making the final predictions. 

\emph{Static-OP} is a supervised gradient boost model \cite{gbm} whose overall structure is given in Algorithm \ref{alg:staticop}. The data used to train \emph{Static-OP} comprises static features and outcomes of the same patients used to construct $\boldsymbol{D_{0+1}^d}$. We denote those by $\boldsymbol{D_{0+1}^s}$ and $\boldsymbol{y_{0+1}}$ respectively. The reconstruction loss  $ \boldsymbol{\hat{p}_{Dynamic}}$ generated by \emph{Dynamic-KD} serve as sample weights $\boldsymbol{\omega}$ (line 1) for each sample in $\boldsymbol{D_{0+1}^s}$. Because \emph{Dynamic-KD} ensures that $\boldsymbol{\hat{p}_{Dynamic}}$ creates a separation between positive and negative classes, the minority samples of $\boldsymbol{D_{0+1}^s}$ will be the determinant of the decision threshold the model is trained to discover.

Using the model to predict the outcome probabilities from $\boldsymbol{y^{s-test}}$ produces $\boldsymbol{\hat{p}}$, and will also produce the variable importance vector (line 2). The class labels $\boldsymbol{\hat{y}}$, are obtained as in Equation \ref{eq:yhat} (line 4), using a prediction threshold learned by maximising PR-AUC from the predicted probabilities (line 3).

\begin{algorithm}[ht]
\SetAlgoLined
\SetKwInOut{Input}{Receives} 
\SetKwInOut{Output}{Returns}

\Input{\begin{enumerate}
 \item  $\boldsymbol{D_{0+1}^s}$ and $\boldsymbol{y^s}$: Static-feature dataset of the same patients used to validate \emph{Dynamic-KD} , with corresponding true-outcome labels. 
				\item  $\boldsymbol{D_{0+1}^{s-test}}$ and  $\boldsymbol{y^{s-test}}$:  An independent mixed-class static testing dataset, with corresponding true-outcome labels. 
\item Validation reconstruction loss, $\boldsymbol{\hat{p}_{Dynamic}}$ of $\boldsymbol{D_{0+1}^s}$ obtained from \emph{Dynamic-KD}.  \end{enumerate}}

\Output{ Classification label $\boldsymbol{\hat{y}}$, $\hat{y} \in \{0,1\}, \forall \hat{y} \in \boldsymbol{\hat{y}}$, variable importance $\boldsymbol{\mathcal{I} }$}

\nonl \;
 
\nonl    \textbf{Train Gradient Boost}

$ \mu =$ \small{TrainGB}$(\boldsymbol{X} = \boldsymbol{D_{0+1}^s}, y = \boldsymbol{y^s}, \boldsymbol{\omega} = \boldsymbol{\hat{p}_{Dynamic}})$

  \nonl \textbf{Test Gradient Boost}
  
  	$\boldsymbol{\hat{p}}, \mathcal{I}  = \mu (\boldsymbol{D_{0+1}^{s-test}, y^{s-test}})$
  	
  	$\gamma = \displaystyle\argmax_{PR-AUC} (\boldsymbol{y^{s-test}},\boldsymbol{ \hat{p}})$
  	
  	$\boldsymbol{\hat{y}}   \overset{\gamma}{\geq}{ \boldsymbol{\hat{p}}}$  
 \caption{Static-OP}\label{alg:staticop}
\end{algorithm}

\section{Experimental Evaluation on Real Use Cases}

We critically assess the model's performance using two sources of data. The first is general-ward (non-ICU) hospital data obtained from King's College Hospital and Princess Royal University Hospital in London, UK. The compiled dataset comprised inpatients diagnosed with COVID-19 for which \emph{KD-OP} was used to predict (a) mortality and (b) ICU admission. The second source is an ICU database, namely the freely-available Medical Information Mart for Intensive Care III (MIMIC-III) \cite{mimic}. Here, we extracted two datasets corresponding to Pneumonia and Chronic Kidney Disease (CKD) patients. Since MIMIC-III strictly contains ICU data, \emph{KD-OP} was used to predict (a) mortality and (b) ICU re-admission from the MIMIC-III cohorts. \emph{KD-OP} was trained to predict adverse outcomes using time-series corresponding to the first 24 hours of hospital admission (in the COVID-19 use case) or ICU admission (in the Pneumonia and CKD use cases). Moreover, the risk of adverse outcomes (mortality, ICU admission or ICU re-admission) was predicted at intervals of 5, 7, 14 and 30 days within admission. 

\subsection{Datasets }\label{sec:datasets}

\subsubsection{COVID-19 Case Study } Data was collected from 1,276 adult ($\geq 18$ years old) inpatients  testing positive for severe acute respiratory syndrome coronavirus 2 (COVID-19) by reverse transcription polymerase chain reaction (RT-PCR) between the $1^{st}$ of March and $31^{st}$ April 2020 at two acute hospitals (King's College Hospital and Princess Royal University Hospital) in South East London (UK). Only symptomatic patients who required inpatient admission were included. Presenting symptoms included but were not limited to fever, cough, dyspnoea, myalgia, chest pain, or delirium.
 Static data collected include age, sex, ethnic background, the length of the period from symptoms onset to hospital admission, and pre-existing conditions (specifically, chronic obstructive pulmonary disease (COPD), Asthma, heart failure, diabetes, ischemic heart disease (IHD), hypertension and chronic kidney disease). For training and risk prediction, pre-existing conditions were aggregated into one ordinal feature describing the number of comorbidities at the time of admission. The dynamic features included 14 of the routinely collected vital signs and laboratory tests of Table \ref{tab:aggregation}. Details of the variables used in the COVID-19 study and their missingness ratios in the data collected are available on our online repository
 \footnote{https://github.com/zibrahim/MIMICTimeSeriesProcessing /blob/main/VitalAggregation.pdf}.

\begin{center}
\begin{table}[ht]
\resizebox{\columnwidth}{!}{%
\begin{tabular}{ l l l l } 
 \hline
    \textbf{Attribute} & \textbf{COVID-19} & \textbf{Pneumonia} & \textbf{CKD} \\
        \textbf{} & \textbf{(General Ward)} & \textbf{ICU}  & \textbf{(ICU)}\\\hline

    \textbf{Patients, n} & 1,276 & 2,798 & 2,822\\
    \textbf{Females} &  735(57.6 \%) & 1,217(45.58\%) & 1,139 (37.27\%)\\
    \textbf{Age} & 69.3 (16.79) & 50.05 (33.93) & 72.61 (22.55)\\
    \textbf{No. Comorbidities} &  0.32 (0.14)  & 0.22 (0.18) & 0.29 (0.19)\\
    
    \textbf{Mortality} &&  & \\
           \multicolumn{1}{r}{\textbf{5-Day}}  &   139 (10.88\%) & 183 (6.48\%)  & 169 (5.98\%)\\
       \multicolumn{1}{r}{\textbf{7-Day}}  &   187(14.64\%) & 257 (9.19\%) & 222 (7.87\%)\\
    \multicolumn{1}{r}{\textbf{14-Day}}  & 264 (20.68)  &  412 (14.73\%)  & 310 (10.98\%) \\
    \multicolumn{1}{r}{\textbf{30-Day}}  & 335 (26.23\%) &   530 (18.94\%)  & 378 (13.39\%)\\
        \textbf{Days to mortality} & 12.07 (15.59) &  15.38 (12.46) & 11.87 (14.44)\\
        \textbf{ICU} & 12.07 (15.59) &  15.38 (12.46) & 11.87 (14.44)\\

           \multicolumn{1}{r}{\textbf{5-Day}}  &   105 (8.22 \%) & 63 (2.36\%)  & 78 (2.81\%)\\
       \multicolumn{1}{r}{\textbf{7-Day}}  &   112 (8.77\%) & 102 (3.82\%)  & 80 (2.84\%)\\
    \multicolumn{1}{r}{\textbf{14-Day}}  & 123 (9.63\%)  &  199 (7.45\%)  & 157 (5.56\%)\\
    \multicolumn{1}{r}{\textbf{30-Day}}  & 124 (9.71\%) &   272 (10.18\%)  & 196 (6.95\%)\\
            \textbf{Days to ICU} & 4.35(8.30) &  12.23 (10.46) & 11.18 (10.49)\\

 \hline
\end{tabular}}
\caption{\label{tab:cohorts} Summary statistics of the datasets. Data are displayed as mean (standard deviation) or count (percent). The number of pre-existing conditions was normalised to [0,1] as the two data sources used different scales to measure co-morbidity. ICU refers to ICU admission in the COVID-19 study and ICU re-admission In MIMIC-III. }
\end{table}
\end{center}


    
     

\subsubsection{Pneumonia and CKD Case Studies} 
We used the data of ICU stays between 2001 and 2012 obtained from the MIMIC-III database, which is a freely-available anonymised ICU database and is the largest resource of time-series hospital data available worldwide \cite{mimic}. We extracted admission details, demographics, time-stamped vital signs and laboratory test results obtained over the first 24 hours of admission of adults having ICD-9 code = 482.9 (pneumonia, cause not otherwise specified) and 585.9 (CKD, cause not otherwise specified) as the primary diagnoses in the ICU admission notes. Our choice of relying on ICD-9 codes to construct the two cohorts is a pragmatic one. ICD-9 codes are highly-specific but exhibit low sensitivity in extracting confirmed diagnoses from EHRs, including those of pneumonia \cite{pneumonia}, and CKD \cite{ckd}. Since our aim is to evaluate the developed model, ICD-9 codes are sufficient to extract correct yet possibly incomplete cohorts for the two conditions. 

Since the MIMIC-III database is structured such as each hospital admission may correspond to multiple ICU stays, we extract the time-series pertaining to the first ICU stay of each admission, and used subsequent ICU admission to ascertain readmission outcomes. The resulting datasets comprise 509,323 records corresponding to 2,798 pneumonia ICU stays and 702,813 records corresponding to 2,822 CKD ICU stays (SQL and python scripts for recreating the dataset using the MIMICIII database are available on our online repository\footnote{https://github.com/zibrahim/MIMICTimeSeriesProcessing}). 

\subsubsection{Data Description and Characteristics}

Table \ref{tab:cohorts} provides statistical summaries of the three datasets, including demographics and outcome distributions in the training and test sets of the three cohorts. The datasets vary in size, where pneumonia and CKD are much larger than COVID-19. The difference in size is a direct consequence of the mode of collection. The pneumonia and CKD datasets were extracted from the largest publicly-available ICU time-series database \cite{mimic}, while the COVID-19 data was locally collected over a short time span. Females were only the majority of cases in the COVID-19 dataset (females = 57.6 \%), but sex distribution only significantly differed from the CKD dataset (females = 37.27\%) and not from the pneumonia dataset (females = 45.58\%). The pneumonia cohort was significantly younger and less co-morbid than the other two. The pneumonia cohort also showed a wider distribution of age compared to COVID-19 and CKD . In addition, the table shows that the number of pre-existing conditions varied greatly in the pneumonia and CKD cohorts, while the distribution of pre-existing conditions was more uniform in the COVID-19 dataset.  The different distributions in age and pre-existing conditions is quite reasonable and align with the nature of the use cases: CKD is an age-related chronic illness \cite{bloodpressureage}, with previous studies showing that the rate of comorbidities is around 41\% \cite{ckdmm}. COVID-19 hospital admissions are more likely in the elderly with pre-existing conditions such as hypertension and diabetes, where symptoms are likely to be more severe as opposed to the young  healthy individuals \cite{covidold}. In contrast, although both older age and pre-existing conditions increase the risk of acquiring pneumonia, they have not been found to be associated with the severity of the condition and subsequent intensive care needs \cite{pneumoniaage,pneumonia2}. 

Across all prediction intervals, the COVID-19 dataset had higher rates of mortality, while mortality rates of CKD were significantly lower than in the other two cohorts. In addition, the CKD cohort had significantly lower rates of ICU re-admissions across all prediction intervals compared pneumonia, and compared to ICU admission in COVID-19. However, the time to ICU admission in COVID-19 was much lower than the time to ICU re-admission in CKD and pneumonia, where the average duration was 4.35 days, compared to 12.23 days in pneumonia and 11.18 days in CKD. In all three cohorts, the training and test sets are comparable in age, percentage of females and the distribution of outcomes (5,7, 14 and 30-day mortality, ICU admission or re-admission).


\subsection{Implementation Details}

Applying the regularisation procedure described in Section \ref{sec:processing} produced $T= 48$ for pneumonia and CKD (aggregation into half-hourly windows) and $T=12$ for the COVID-19 use-case (aggregation into 2-hourly windows) for the 24-hour data extracted. The resulting datasets comprised 30,624 samples with 12 variables for the COVID-19 dataset, 134,304 and 135,456 samples with 30 variables for the pneumonia and CKD cases respectively.

\begin{figure}[ht]

\centering
\includegraphics[width=.5\textwidth]{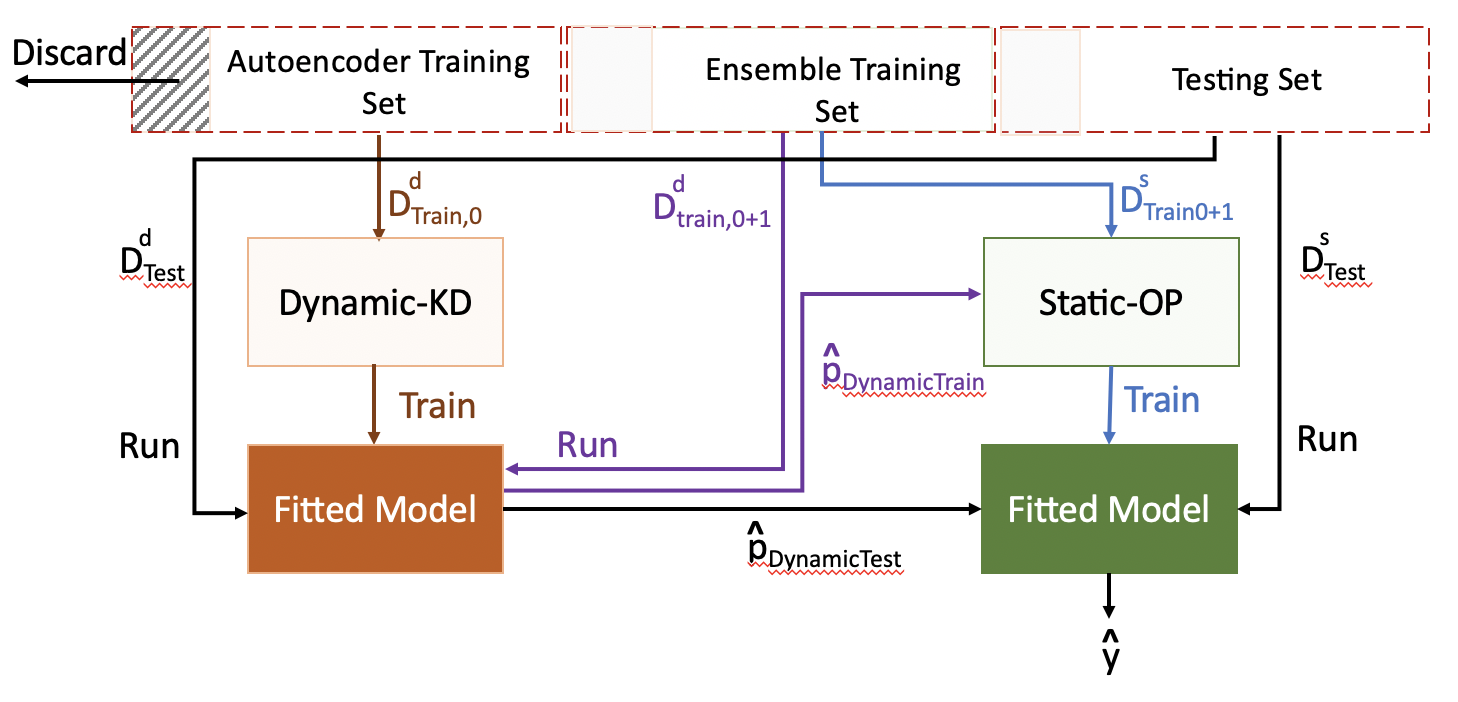}

\caption{The procedure followed for the three-way split of the data in \emph{KD-OP}. Each group shows an approximation of the proportion of positive-outcome (shaded) to negative-outcome samples (non-shaded). The positive outcomes have been marked as discarded in the autoencoder training set because the unsupervised \emph{Dynamic-KD} is trained solely on positive-class time series. The positive class samples of the ensemble training and testing sets have been highlighted for visualisation purposes only.}
\label{fig:datasplit}

\end{figure}

The datasets have two notable properties: 1) the outcomes are skewed, with positive outcomes being highly under-represented in the time series, and 2) the temporal ordering is defined over $T$ batches for each patient. In order to retain the natural distribution of outcomes and temporal ordering during training and to prevent information leakage, we used stratified grouped k-fold cross-validation\footnote{https://www.kaggle.com/jakubwasikowski/stratified-group-k-fold-cross-validation}, with k=3, to split the data as shown in Figure \ref{fig:datasplit}. At each iteration, the data used to train \emph{Dynamic-KD} is obtained using one fold (Autoencoder Training Set in the figure), discarding the samples corresponding to patients with positive outcomes to yield $\boldsymbol {D^d_{Train,0}}$. The second fold (Ensemble Training Set in the figure) is used to run \emph{Dynamic-KD} to predict $\boldsymbol {\hat{p}_{DynamicTrain}}$ and further to train \emph{Static-OP}, using  $\boldsymbol{\hat{p}_{DynamicTrain}}$ as sample weights. The third fold (Testing Set in the figure) is used to predict the testing $\boldsymbol {\hat{p}_{DynamicTest}}$ using \emph{Dynamic-KD}, using it as sample weights in \emph{Static-OP}, where the final model prediction is made to produce $\boldsymbol{\hat{y}}$.

\begin{figure}[hbt!]
 \centering
\includegraphics[width=.35\textwidth,keepaspectratio]{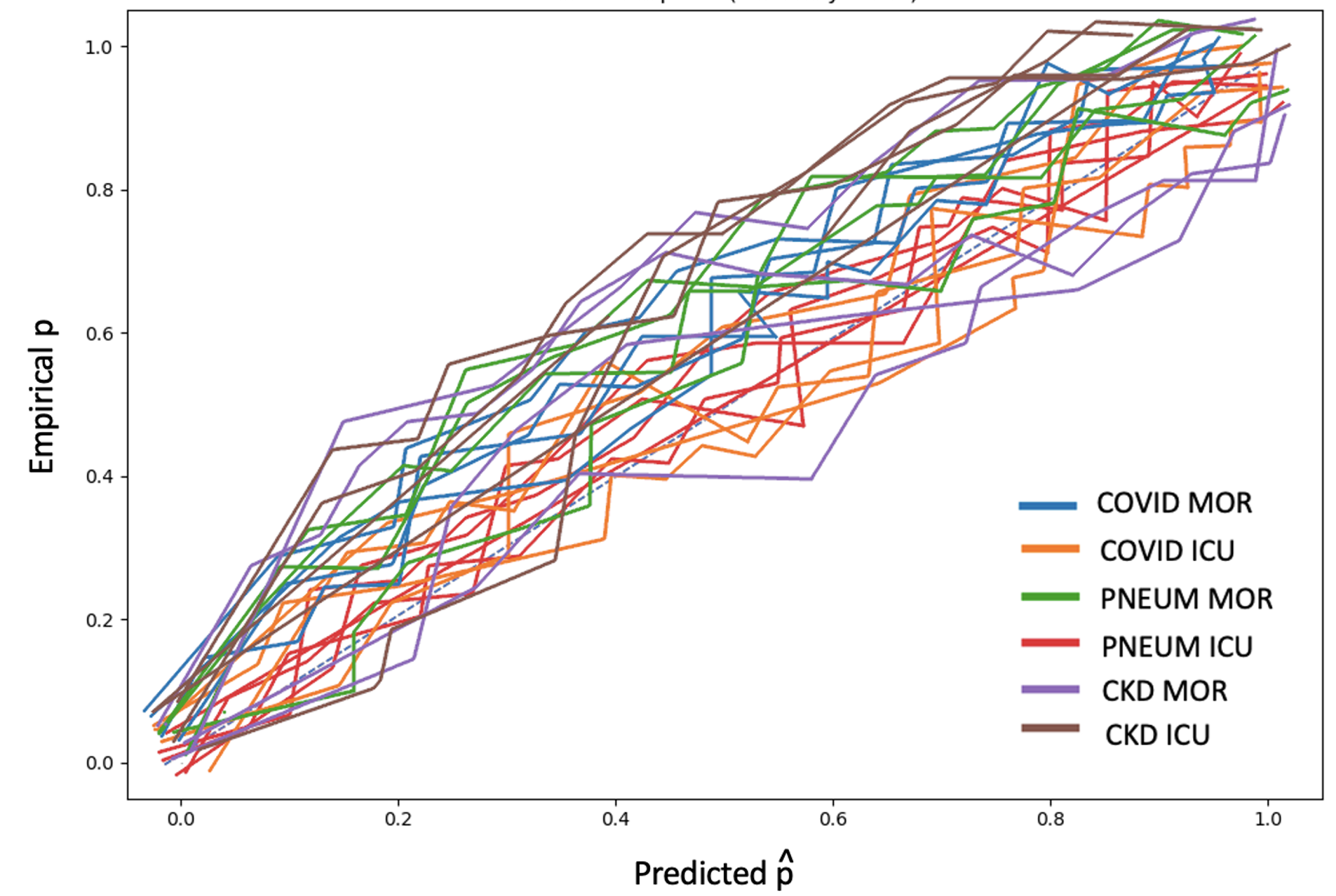}\caption{\label{fig:reliability} Reliability plot of \emph{KD-OP}'s predicted $\boldsymbol{\hat{p}}$ in the test set, where empirical $\boldsymbol{p}$ is the fraction of patients with adverse events within the interval (e.g. 30-day mortality).}
\end{figure}

We used the Python language and the Keras library with Tensorflow backend\footnote{The source is available at: https://github.com/zibrahim/StackedPredictor/}. For \emph{Dynamic-KD}, model hyperparameters were optimised through empirical evaluation, by carefully observing the prediction performance using a set of candidate values of the  hyperparameters; those included the number of neurons per layer, the number of hidden layers, dropout rates and activation function. The final design included bi-layered encoder and decoder, with the outmost layers having neurons in the order of $2\times n\_features$, where $n\_features$ is the number of dynamic variables used (14 for COVID-19 and 30 in the pneumonia and CKD studies). A dropout rate of 0.5 was used between each two layers to prevent the autonecoder from overfitting the training data and an adaptive learning rate was used using the Adam optimizer and starting at 0.001. The number of epochs was 1,000, which was selected via cycles of experiments and careful monitoring of the validation loss. An early stopping criteria was used to retain the best model by minimising the validation loss with a patience parameter of 50 epochs. All layers of the autoencoder used ReLU as their activation function, which performed best during our evaluation. The \emph{Static-OP} module was implemented using the XGBoost algorithm. The parameters were chosen through a grid-search over the hyperparameter space. \emph{Static-OP}'s sample weights were set to \emph{Dynamic-KD}'s prediction loss. We used calibrated classification using the nonparametric isotonic method included in Python's scikit-learn package. The calibration plot is shown in Figure \ref{fig:reliability}. 

\subsection{Results}

We evaluate \emph{KD-OP}'s performance across four dimensions. First, we evaluate the performance under different settings presented by the three datasets; these include cohort heterogeneity with respect to individual characteristics and outcome distribution with respect to the minority (positive) cases. Here, we initially report metrics averaged across the different prediction intervals for each setting to obtain an overall view, and subsequently evaluate the model's robustness across different prediction intervals. We then evaluate the contributions of the two modules \emph{Dynamic-KD} and \emph{Static-OP} to the overall performance, validating those empirically and against clinical knowledge. Finally, we compare the predictive power of \emph{KD-OP} with existing outcome prediction models as reported in the literature. After evaluating \emph{KD-OP}'s performance, we demonstrate its visualisation capability in section \ref{sec:justfification}.

Throughout the experiments, we report the Precision-Recall Area Under the Curve (PR-AUC) to capture the model's performance with respect to the minority cases, as well as the widely-used Receiver-Operator Area Under the Curve (ROC-AUC). Despite our knowledge of ROC-AUC's impartial assessment of the model's performance for positive and negative outcomes \cite{imbalanced}, we choose to show it here due to its wide usage in the literature. Specifically, we use ROC-AUC to compare our model's performance with state-of-the-art models in section \ref{sec:comparelit}. We also report the macro-averaged precision, recall and F1-score. We used macro averages to understand the modules' true performance with respect to the under-represented outcomes \cite{imbalancedbook2}. 

\subsubsection{Overall Performance and Sample Diversity}\label{sec:diversity}

\begin{figure}[ht!]

\centering
  \texttt{COVID-19} 
    \bigskip

    \begin{subfigure}[t]{.45\linewidth}
\centering\includegraphics[width=\linewidth]{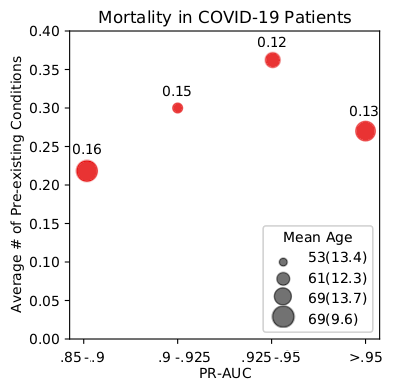}\caption{Mortality}
\end{subfigure}
\begin{subfigure}[t]{.45\linewidth}
\centering\includegraphics[width=\linewidth]{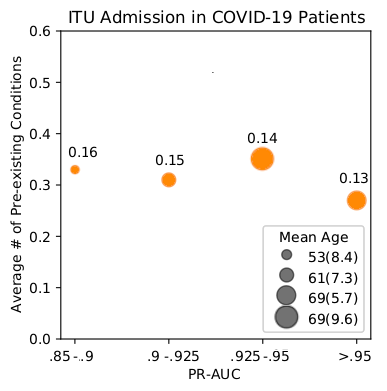}\caption{ICU Admission}
\end{subfigure}

\texttt{Pneumonia}

\bigskip

\begin{subfigure}[t]{.45\linewidth}
\centering\includegraphics[width=\linewidth]{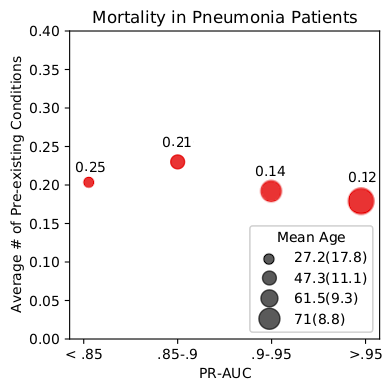}\caption{Mortality}
\end{subfigure}
  \bigskip
    \begin{subfigure}[t]{.45\linewidth}
\centering\includegraphics[width=\linewidth]{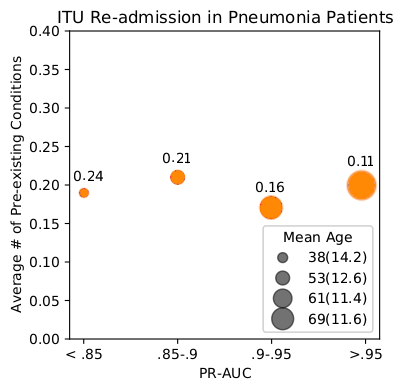}\caption{ICU Re-admission}
\end{subfigure}

\texttt{CKD}

\bigskip

\begin{subfigure}[t]{.45\linewidth}
\centering\includegraphics[width=\linewidth]{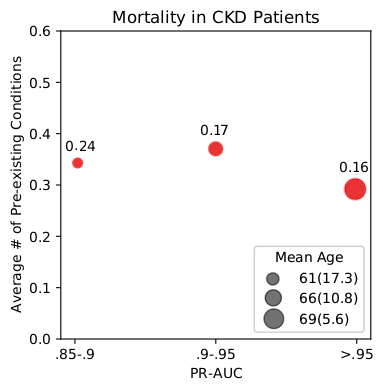}\caption{Mortality}
\end{subfigure}
\begin{subfigure}[t]{.45\linewidth}
\centering\includegraphics[width=\linewidth]{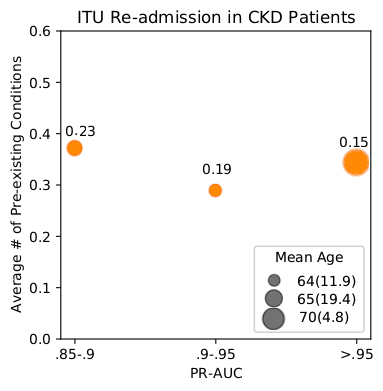}\caption{ICU Re-admission}
\end{subfigure}

\caption{Grouped performance plot. The x-axis corresponds to performance grouped by PR-AUC. The mean comorbidity per performance group is given by the y-coordinates (circles), while the comorbidity standard deviation is shown as text annotations over each circle. The circle size indicates the mean age within the performance range, while the numerical age means and standard deviations are shown in the legend.}\label{fig:het}

\end{figure}

We first evaluate the overall performance across the three case studies. For each dataset, Table \ref{tab:overallperformance} shows the model's performance averaged across the prediction intervals of 5, 7, 14 and 30 days for each outcome. The performance is high overall. However, better performance across prediction intervals was achieved using the COVID-19 dataset compared to pneumonia and CKD, despite the latter two being larger datasets with a higher resolution of observations (half-hourly windows as opposed to two-hourly windows used to construct the COVID-19 time-series). A close examination is shown in Figure \ref{fig:het}, where higher performance ranges appear to be closely correlated with sample homogeneity (lower standard deviation) for both age and the number of pre-existing conditions. COVID-19 admissions are more uniform in age and comorbidity, consequently influencing changes in physiological states. In pneumonia, the standard deviation of comorbidity is higher in younger patients where the model achieves the lowest performance. 

\begin{center}
\begin{small}
\begin{table*}[ht!]
\resizebox{\textwidth}{!}{%
\begin{tabular}{l c c  | c c | c c } 
 \hline
&    \multicolumn{2}{c}{  \textbf{COVID-19} }&\multicolumn{2}{c}{ \textbf{Pneumonia}} & \multicolumn{2}{c}{  \textbf{CKD} }\\\hline
	& \textbf{Mortality} & \textbf{ITU Admission} & \textbf{Mortality }&\textbf{ ITU Readmission}& \textbf{Mortality }&\textbf{ ITU Readmission}\\	
\textbf{Avg Macro Prec} & 0.918 (0.815-0.931)& 0.939 (0.902-0.946)& 0.881 (0.814-0.902)  & 0.873 (0.737-0.931)  & 0.916 (0.894-0.921)& 0.868 (0.760-0.905) \\
\textbf{Avg Macro Rec} & 0.938 (0.907-0.969)& 0.970 (0.963-0.984) &  0.884 (0.865-0.892)&  0.869 (0.825-0.926)& 0.953 (0.853-0.906) & 0.889 (0.802-0.985)\\
\textbf{Avg Macro F1 } &0.936 (0.913-0.963)& 0.977 (0.976-0.978) &   0.901 (0.90-0.984)& 0.881 (0.774-0.951)& 0.912 (0.883-0.935)&  0.919 (0.781-0.985)\\
\textbf{Avg PR-AUC} &  0.922 (0.881-0.959)& 0.948 (0.851-0.965) & 0.871 (0.830-0.923)&  0.881 (0.727-0.967)  & 0.896 (0.869-0.938)& 0.895 (0.756-0.998)\\
\textbf{Avg ROC-AUC }& 0.931 (0.896-0.942) & 0.971 (0.969-0.987) & 0.884 (0.865-0.892)& 0.870 (0.825-0.926) &0.878 (0.853-0.908)& 0.895 (0.826-0.945)\\

 \hline
\end{tabular} }

\caption{\label{tab:overallperformance}Average performance per outcome for COVID-19, pneumonia and CKD. }
\end{table*}
\end{small}
\end{center}

\subsubsection{Performance Across Prediction Intervals and Outcome Distribution Settings}

\begin{figure}[ht]

\centering
  \texttt{COVID-19} 
    \bigskip

    \begin{subfigure}[t]{.45\linewidth}
\centering\includegraphics[width=\linewidth]{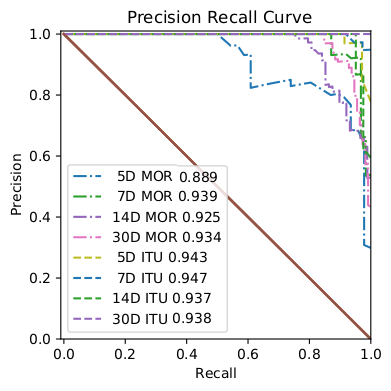}\caption{PR-AUC}
\end{subfigure}
\begin{subfigure}[t]{.45\linewidth}
\centering\includegraphics[width=\linewidth]{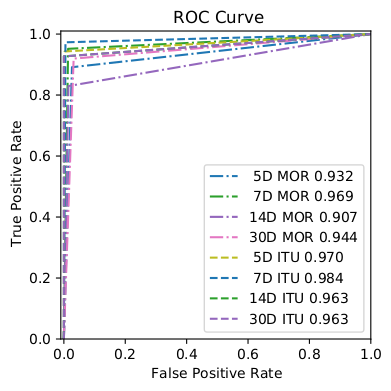}\caption{ROC-AUC}
\end{subfigure}

  \bigskip

\texttt{Pneumonia}
\bigskip

\begin{subfigure}[t]{.45\linewidth}
\centering\includegraphics[width=\linewidth]{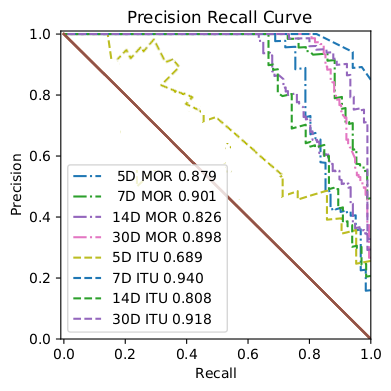}\caption{PR-AUC}
\end{subfigure}
\begin{subfigure}[t]{.45\linewidth}
\centering\includegraphics[width=\linewidth]{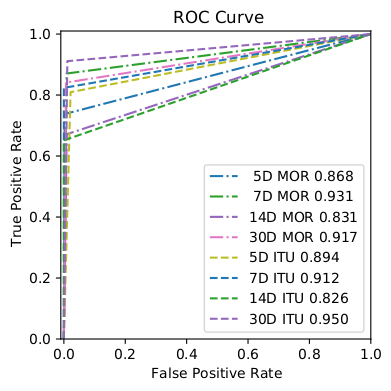}\caption{ROC-AUC}
\end{subfigure}

\centering
  \texttt{CKD} 
    \bigskip

\begin{subfigure}[t]{.45\linewidth}
\centering\includegraphics[width=\linewidth]{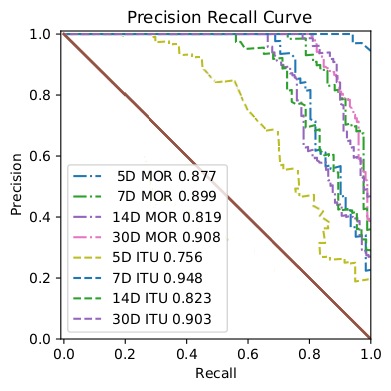}\caption{PR-AUC}
\end{subfigure}
\begin{subfigure}[t]{.45\linewidth}
\includegraphics[width=\linewidth]{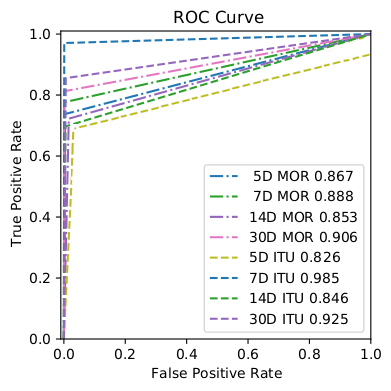} \caption{ROC-AUC}
\end{subfigure}

\caption{Performance measured in PR-AUC (top) and ROC-AUC (bottom) for COVID-19, Pneumonia and CKD. In each plot, the respective performance is shown over 5, 7, 14 and 30 days for mortality and ICU admission (in the COVID-19 case) or ICU re-admission (in the pneumonia and CKD cases). }
\label{fig:performance}

\end{figure}

Figure \ref{fig:performance} shows \emph{KD-OP}'s performance in predicting mortality and ICU admission/re-admission over 5, 7, 14 and 30-day intervals on COVID-19, pneumonia and CKD. Two observations can be made when examining this figure in conjunction with the distribution of the outcomes of Table \ref{tab:cohorts}. Apart from 5-day ICU re-admission in pneumonia and CKD,  (a) \emph{KD-OP} shows high performance across short and long-term intervals, and (b) \emph{KD-OP}'s performance remains high when the minority (positive) samples constitute $< 10\%$ of the overall population, which confirms the merit of relying on outlier detection to construct the temporal representation used in the pipeline. These findings are in-line with the demographic diversity results of Section \ref{sec:diversity}, as the mean and standard deviations of age and the number of pre-existing conditions of those re-admitted to the ICU within 5 days were 33.76 (35.1) and 0.23 (0.2) for pneumonia and  66.12(21.3) and 0.28(0.17) for CKD respectively, showing a highly-diverse sample, deviating almost as highly as the overall pneumonia and CKD populations as shown in table \ref{tab:cohorts}. In contrast, the 5-day ICU admission sample in the COVID-19 study had a mean and standard deviation of age and the number of pre-existing conditions of 63.3 (9.79) and 0.32 (0.05) respectively, showing a narrow range of demographical variation compared to pneumonia and CKD.


\subsubsection{The Contribution of \emph{Static-OP} vs \emph{Dynamic-KD}}
We now turn to compare the relative contribution of the two modules to \emph{KD-OP}'s overall prediction across the three use cases, outcomes and four intervals. The detailed comparison is provided in table \ref{tab:performance1}. In the table, we list the contribution of each module per prediction interval for each outcome using macro-averaged precision, recall and F1-score, as well as PR-AUC and ROC-AUC. We also show \textbf{avg} $\Delta$, the average change in each metric's value between \emph{Dynamic-KD} and the final prediction made by \emph{KD-OP}. It is clear that the two modules complement each other to reach a high performance that is not otherwise achievable by the time-series predictor alone. This effect is especially noticeable in recall, where \emph{Static-OP} often increases \textbf{avg} $\Delta$.  \textbf{avg} $\Delta$ of mortality outcomes also show that the stacked model slightly decreases \emph{Dyanmic-KD}'s precision, indicating that the temporal signatures are somewhat better suited for identifying more patients who are at risk of mortality. However, the magnitude of the decrease in the overall precision is insignificant compared to the magnitude of increased recall, leading to a higher decrease in false alarms, which is a common bottleneck in clinical outcome prediction models \cite{news2performancelalala}. 


Examining the performance from a domain angle, \emph{Static-OP}'s contribution to the overall performance appears to be more pronounced in short-term outcomes. A highly noticeable difference is in the case of COVID-19 5-day mortality, where the average macro F1-score increases by 0.127 (from 0.786 using \emph{Dynamic-KD} alone to 0.913 using the full pipeline). In contrast, the increase in F1 goes down to 0.021 (from 0.922 to 0.943) when examining 30-day mortality. This observation is consistent with current knowledge and recent findings that demographic information (e.g. age, pre-existing conditions) are highly predictive of short-term mortality in COVID-19 patients \cite{demographics}. Similarly, for ICU readmission, replicated studies have found co-morbidities to be highly predictive of intensive care readmission during the same hospitalization \cite{readmission}.

\begin{table}[ht]
\begin{center}
\resizebox{\columnwidth}{!}{%
\begin{tabular}{ l l l c c c c c}

\hline \textbf{Dataset} &  \textbf{Outcome} & \textbf{Model} & \textbf{Averge} & \textbf{Average} & \textbf{Average }  &  \textbf{PR-AUC} & \textbf{ROC-AUC}\\ 
  \textbf{} & \textbf{} & &  \textbf{Macro} &  \textbf{Macro} &  \textbf{Macro } &\\ 
    \textbf{} & \textbf{} & &  \textbf{Pr} &  \textbf{Rec} &  \textbf{F1 } &\\  \hline 

  \textbf{General}&   5MOR & Dyn-KD &0.917 &0.717& 0.786  &  0.553 &  0.683 \\
  \textbf{(COVID)}& & KD-OP&  0.865 & 0.932& 0.913 & 0.889 & 0.932 \\ \hline
   					& 7MOR & Dyn-KD & 0.931 & 0.882 & 0.927 & 0.855 & 0.887 \\
   					& 				& KD-OP 		  & 0.922 & 0.984 & 0.978 &0.939 &  0.969  \\ \hline
   					& 14MOR & Dyn-KD & 0.898 & 0.824 & 0.871 &   0.782 & 0.823 \\
   					& 					& KD-OP & 0.896 & 0.907 & 0.925 &  0.925 &  0.907\\\hline
   					& 30MOR & Dyn-KD & 0.915 & 0.893 & 0.922 &  0.885 & 0.895  \\
   					& 					& KD-OP & 0.899 & 0.944 & 0.943 &  0.934  & 0.944 \\\hline
				   	&					 & \textbf{Avg} $\boldsymbol{\Delta}$ 	&  \textbf{-0.019 }&\textbf{+0.128} &  \textbf{+0.065}&  \textbf{+0.153} & \textbf{+0.119}\\\hline

   					& 5ICU 		& Dyn-KD & 0.918 & 0.897 & 0.925 & 0.853 & 0.932 \\
   					& 					& KD-OP & 0.933 & 0.970 & 0.976 & 0.943 & 0.970 \\\hline
   					& 7ICU 		& Dyn-KD & 0.944 & 0.932 & 0.961 &  0.885 & 0.887 \\
   					& 					& KD-OP & 		0.933 & 0.970 & 0.976 &  0.947 & 0.984 \\\hline
   					& 14ICU 	& Dyn-KD & 0.929 & 0.826 & 0.950 & 0.782 & 0.823 \\
   					& 					& KD-OP & 0.946 & 0.963 & 0.979 &  0.937 & 0.963 \\\hline
   					& 30ICU 	& Dyn-KD & 0.933 & 0.962 & 0.972 & 0.994 & 0.999 \\
   					& 					& KD-OP & 0.946 & 0.963 & 0.979 & 0.938  & 0.963  \\\hline
				   	&					 &	 \textbf{Avg} $\boldsymbol{\Delta}$ 	& \textbf{+0.008}& \textbf{+0.245}& \textbf{+0.026} & \textbf{+0.063}\ & \textbf{+0.059}\\\hline
					
 \textbf{ICU}&	 5MOR	& Dyn-KD  & 	0.938	& 0.736		& 0.896		&   0.72  & 0.8 \\
 \textbf{(Pneum)}	& 					& KD-OP & 0.904 & 0.891 & 0.919 &0.879 & 0.868 \\\hline
    						& 7MOR & Dyn-KD & 0.934 & 0.841 & 0.997 & 0.765 & 0.862 \\
    						& 				& KD-OP & 0.911 & 0.892 & 0.923  & 0.901 & 0.931 \\\hline
    						& 14MOR & Dyn-KD & 0.934 & 0.841 & 0.897  & 0.73 & 0.802 \\
    						& 				  & KD-OP & 0.894 & 0.865 & 0.90 &  0.826 & 0.831\\\hline
    						& 	30MOR			& Dyn-KD  & 0.922 & 0.878 & 0.916  & 0.861 & 0.885 \\
    						& 							& KD-OP & 0.917 & 0.888 & 0.921& 0.898 &0.894\\\hline
					   	&					 & \textbf{Avg} $\boldsymbol{\Delta}$ 	& \textbf{-0.026} & \textbf{+0.035}& \textbf{-0.011}& \textbf{+0.107} & \textbf{+0.044}\\\hline
    						
    						& 5ICU 				& Dyn-KD & 0.648 &	0.684 & 0.679 &  0.587 & 0.643 \\
    						&							& KD-OP &  0.737 & .825& 0.774& 0.689 & 0.894 \\\hline
    						& 7ICU 				& Dyn-KD & 0.944& 0.853 & 0.999 &0.967 & 0.926 \\
    						& 							&  KD-OP	& 0.931	& 0.926& 0.951&   0.940 & 0.912 \\\hline
    						& 14ICU				& Dyn-KD & 0.950	& 0.742& 0.816& 0.561 & 0.713 \\
    						& 							& KD-OP & 0.894 &	0.835& 0.866&  0.808 & 0.826 \\\hline
    						& 	 30ICU 			& Dyn-KD  	& 0.816&	0.818 & 0.899& 0.764 &  0.828 \\
    						& 							& KD-OP & 0.931	& 0.894 & 0.932  & 0.918 &  0.950 \\\hline
    						&					 &\textbf{Avg} $\boldsymbol{\Delta}$ 	& \textbf{+0.034}& \textbf{+0.096}& \textbf{+0.032} & \textbf{+0.119} & \textbf{+0.086}\\\hline
\textbf{ICU} 	& 5MOR & Dyn-KD & 0.938     & 		0.836	& 0.896		&  	0.766	& 	 0.799\\
\textbf{(CKD)}& 					& KD-OP			& 	0.909	& 	0.897		& 	0.915		& 	0.887			& 	0.867	\\\hline

						& 7MOR	& Dyn-KD & 0.934 & 0.841 		& 0.915 	&	0.728		& 		0.855		\\
                        & 						& KD-OP 		& 0.931 & 0.901 & 0.947				& 	0.899	&   0.888\\\hline
                        & 14MOR	& Dyn-KD 	& 0.917 & 0.807 	& 0.863 & 0.728		& 		0.806 		\\
                        & 						& KD-OP 		& 0.909 & 0.863 & 0.902 & 	0.819	& 	0.853	\\\hline
                        & 30MOR & Dyn-KD 	& 0.918 & 0.879 & 0.916 &	0.862		 & 		0.903	\\
                        & 					& KD-OP &			 0.924 & 0.906 & 0.935 & 		0.908		& 	0.906	\\\hline
                       	&					 & \textbf{Avg} $\boldsymbol{\Delta}$ 	&  \textbf{-0.009 }&\textbf{+0.050} &  \textbf{+0.027} & \textbf{+0.107} & \textbf{+0.0375}\\\hline

                        & 5ICU		& Dyn-KD 	& 0.612 & 0.646 & 0.665 &  	0.658	& 		0.733	\\
                        & 					& KD-OP & 0.727 & 0.772 & 0.739 &   0.756	& 	0.826	\\\hline
                        &  7ICU		& Dyn-KD & 0.944 & 0.853 & 0.911 &0.778 & 0.890 \\
                        &  					& KD-OP & 		0.935 & 0.985 & 0.985& 0.948 & 0.985 \\\hline
                       &    14ICU & Dyn-KD & 0.898 & 0.743 & 0.816 &  0.705 & 0.714 \\
                          & 					& KD-OP &0.898 & 0.846  & 0.920 &  0.823 & 0.846\\\hline
                         &   30ICU &Dyn-KD & 0.934 & 0.861 & 0.941 & 0.785 & 0.859 \\
                           	& 				& KD-OP & 0.912 & 0.925 & 0.942& 0.903 & 0.925 \\\hline
                       & & \textbf{Avg} $\boldsymbol{\Delta}$ 	&  \textbf{+0.027}&\textbf{+0.106} &  \textbf{+0.064} & \textbf{+0.126} & \textbf{+0.095}\\\hline
\end{tabular}}
\end{center}
\caption{\label{tab:performance1} Comparison of Dynamic-KD's predictions with the final prediction made by KD-OP across the case studies, grouped by prediction interval (5, 7, 14 and 30 days) and by outcome (mortality, ICU admission/re-admission.}
\end{table}



\subsubsection{Comparison with Existing Outcome Prediction Models}\label{sec:comparelit}

\begin{table*}
\begin{center}
\resizebox{\textwidth}{!}{%
\begin{tabular}{l l l  l c c c c} 

\hline
 \textbf{Model} &   \textbf{Primary Outcome} &\textbf{Outcome Distribution }&  \textbf{AUC} & \textbf{ PR-AUC}  & \textbf{Sensitivity} & \textbf{Specificity}  \\ \hline
 \textbf{General Ward} & \\
 NEWS2 \cite{news2performancelalala} & Deterioration (1 day) &  (94.8\%, 5.2\% )& 0.78 (0.73 - 0.83) & NA  & 0.28 (0.21 - 0.37) & 0.80 (0.79 0.82)\\
 eCART \cite{mortalityprediction2,mortalitypredictionlatest} & Unplanned ICU Admission, & (95\%, 5\%) &0.75 (0.74–0.75) & NA  & NA & NA \\
 &  Mortality     & (98.8\%, 1.2\%)  &  0.93 (0.93–0.93)& NA  & NA  &NA  \\
 & Cardiac Arrest &   (99.5\%, 0.05\%)   &      0.89 (0.88–0.91)    &        & 0.89 (0.88–0.91)&	0.52 (0.52–0.52)\\
 DEWS \cite{attention3shamout} & Unplanned ICU Admission &(72.8\%, 27.2\% ) &  0.811 (0.811 - 0.822) & NA & 0.555 (0.554 - 0.557) & 0.90 (0.90. - 0.90) \\
& Mortality &    (35.7\%, 65.3\%) &   0.926 (0.926 - 0.927)  & NA  &  0.831 (0.831 - 0.832) & 0.888 (0.888 - 0.888)  \\
   LightGBM \cite{brazilian} & Mortality & (94.9\%,  5.1\%)  & 0.961 (NA) & NA & NA & 0.641 \\
  KD-OP (General Ward) &  \textbf{Mortality }&  (82.9\%, 18.1\%) & 0.927 (0.9 - 0.932)  & 0.922 (0.881 - 0.959) & 0.905 (0.89 - 0.91) & 0.94 (0.935 - 0.945) \\
  									& \textbf{Unplanned ICU Admission} & (91.19\%, 8.81\%)  &\textbf{0.981 (0.98  - 0.984)} & 0.948 (0.903 - 0.962) & 0.966 (0.901-0.980) & 0.914 (0.902 -0.934)   \\\hline
\textbf{ICU Settings } & \\
SANMF \cite{icumortality} &  Mortality &  (89.98\%, 10.02\%) & 0.848  (0.846, 8.84) & \\
SICULA  \cite{super} & Mortality &  (88.76\%, 12.24\%) & 0.88 (0.87–0.89) & NA & NA & NA \\
 KD-OP (ICU) &\textbf{Mortality} & (89.05\%, 10.95\%) & \textbf{0.881 (0.880 - 0.885)} & 0.8841 (0.841 - 0.913) & 0.953 (0.903 - 0.969) & 0.918 (0.88 - 0.931) \\\hline
 
\textbf{COVID-19}  &\textbf{} & & & & & & \\
 Carr \cite{carr} & Deterioration (14 days) &  -  & 0.78 (0.74 - 0.82) \\
 Guo \cite{guo} & Deterioration (14 days) &  - &   0.67 (0.61 - 0.73) \\
 Liu \cite{zhang} &Mortality \& Deterioration &  -  &   0.74 (0.69 - 0.79) \\
 Galloway \cite{galloway}  & Deterioration & - & 0.72 (0.68 - 0.77)  &  &  & \\
 Gong \cite{gong} &Deterioration  &   - & 0.853 (0.790- 0.916) \\
 
\end{tabular}}
\end{center}
\caption{\label{tab:comparewithliterature} Comparison of KD-OP's performance with existing literature of general outcome prediction tested in general wards and ICU settings. Additionally, a comparison with current statistical models of to predict deterioration in COVID-19 is given. }
\end{table*}

Here, we compare \emph{KD-OP}'s performance with the reported performance of relevant adverse outcome prediction models, showing the results in Table \ref{tab:comparewithliterature}. We restrict the comparison to models that a) predict adverse outcomes (e.g. deterioration, mortality, re-admission, cardiac arrest), and b) report performance exceeding the NEWS2 baseline of ROC-AUC=0.78. Having gone through the literature, the only machine learning frameworks found to have been validated in non-ICU settings were DEWS \cite{attention3shamout}, eCART \cite{mortalityprediction2,mortalitypredictionlatest}, and LightGBM \cite{brazilian} so we list those first. As the NEWS2 score is widely used to predict deterioration, we include the latest evaluation of its performance (from \cite{news2performancelalala}) in the table for comparison. For these models, we compare their performance against the average performance of \emph{KD-OP} when applied to the COVID-19 use-case, as it is a general ward population. For each model, we highlight the class distribution of the target outcome as reported by each model. For \emph{KD-OP}, the class distribution was taken as the average distribution of the outcomes over the intervals evaluated (5, 7, 14 and 30 days), as shown in Table \ref{tab:cohorts}. As none of the models reports PR-AUC, we compiled a list of all reported metrics, using them to compare against \emph{KD-OP}'s performance in conjunction with the widely-used ROC-AUC. 

For mortality, LightGBM achieved a high ROC-AUC of 0.961 with a class distribution of 5.1\% of the outcome. However, the only other available metric for LightGBM is specificity, which is at a low 0.641 and entails a high rate of false alarms generated by the framework. eCart's ROC-AUC is equally distinctive at 0.93 with extremely low frequencies of mortality cases in the data used (1.2 \% of the cases). However, no information on the recall or specificity of the model is available. We, therefore, draw attention the only outcome for which eCart's sensitivity and specificity are investigated, which is cardiac arrest. For this outcome, despite the high ROC-AUC achieved by the model (0.89) at a very low distribution rate of the outcome variable (0.05\%), specificity is at 0.52, which once again shows the over-optimism of the ROC-AUC reported by the model. \emph{KD-OP} achieved a ROC-AUC of 0.978, with mortality averaging at 18.1\% of the samples (ranging between 10.88-26.23\%). DEWS' AUC score was also high at 0.926. However, the class distribution reported in their study is highly skewed in favour of the outcome (65.3\% mortality). With respect to unplanned ICU admissions, \emph{KD-OP} achieved the highest AUC of 0.981 with an average class distribution of 8.81\% (ranging between 8.22 -9.71\%) for the four intervals. DEWS was the closest competitor at 0.811 AUC, albeit with a significantly higher distribution of the positive outcome (27\%). Overall, \emph{KD-OP} shows the highest performance stability across the two outcomes, rendering it a better candidate for general hospitalisation outcome prediction; especially given the lack of thorough assessment of competitive models using metrics suitable for the problem under study. 

We also list high-performing machine learning models that have only been strictly validated in ICU settings; those include SANMF \cite{icumortality}, SICULA  (a.k.a. the super learner) \cite{super} and \cite{mortalitysmall}. It is worth noting that none of the models predicts ICU re-admission. We, therefore, resort to comparing with \emph{KD-OP}'s average performance in predicting mortality when applied to pneumonia and CKD using the MIMIC-III ICU dataset. Also, apart from \cite{mortalitysmall}, which reports sensitivity, the models strictly rely on ROC-AUC to report their performance. We will, therefore, resort to comparing with \emph{KD-OP}'s performance using ROC-AUC. As the table shows, \emph{KD-OP} is the best predictor of mortality in an ICU setting, marginally exceeding SICULA's performance (ROC-AUC of 0.881 vs 0.880). Given that the SICULA's performance is the current benchmark for mortality prediction in the ICU, \emph{KD-OP}'s performance is well-aligned with existing prediction potential. 

Finally, as the literature now contains several statistical models aiming to make prognostic predictions of COVID-19 hospital admissions, we compare those with \emph{KD-OP} applied to the COVID-19 case. It is worth noting that all of the models listed under the COVID-19 section of table \ref{tab:comparewithliterature} are scoring systems aiming to mimic or exceed the performance of NEWS2 in predicting COVID-19 deterioration. Hence, \emph{KD-OP} presents a novel contribution to the COVID-19 use case in being a scalable end-to-end machine learning architecture for predicting hospitalisation outcomes for COVID-19 admissions.

\subsubsection{Visual Justification of Predicted Outcomes} \label{sec:justfification}

   
The stacked nature of \emph{KD-OP} naturally enables visualising its predictions using the built-in visualisation properties of each module and obtaining the relative contributions of each module's prediction to the outcome. For \emph{Dynamic-KD}, the feature attention weights at each time window embedded in $\boldsymbol{C}$  (see output to Visualisation module in Figure \ref{fig:DynamicKD} and line 2 of Algorithm \ref{alg:dynamickd}) make up the relative importance of the temporal signatures of each feature. On the other hand, the gradient boost implementation of \emph{Static-OP} provides a feature importance capability, which we use to understand the relative contribution of each static feature. Since the relative contribution of each module to the final prediction is dependent on the outcome and prediction interval, including those variables in the visualisation of the output is highly essential for clinical utility as it directs the attention to the most contributing view (static or dynamic) of the patient. We define the contribution of each module using the ratios of the respective modules' PR-AUC.

    \begin{figure*}[hbt!]
 \centering
 \includegraphics[width=\textwidth,keepaspectratio]{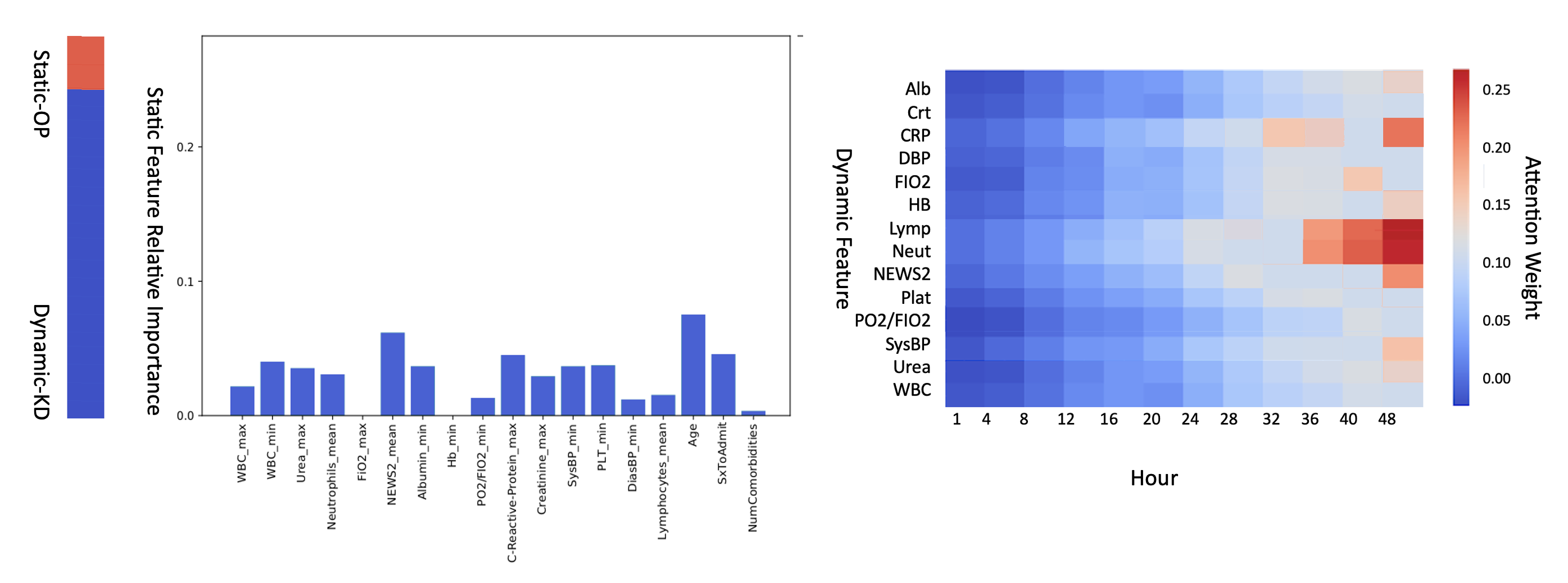}\caption{\label{fig:justification} Visual Justification of KD-OP's Predictions}
\end{figure*}

An example of the generated visualisation of a positive 30-day mortality outcome of a COVID-19 patient is shown in figure \ref{fig:justification}. In the figure, the left-most bar shows the relative contribution of the individual modules. In this scenario, \emph{Dynamic-KD} is a significant contributor ($\approx$ 93\% of the overall contribution). Examining the attention weights generated by the framework shows that the highest weights are of lymphocytes and neutrophils (Lymph and Neut in the figure) at hours 36-48 as well as C-reactive protein (CRP in the figure) at time-step 48 (24$^{th}$ hour). We use this information as a justification for the prediction made by \emph{Dynamic-KD}. On the other hand, the patient's age and the mean NEWS2 score show the highest importance among the static features, followed by the length of the period from symptoms to admission (SxToAdmit) and the maximum C-reactive protein level over the 24 hours. 

It is essential to view this justification in relation to current findings. C-reactive protein, lymphocytes and lactic dehydrogenase have been recently found to be highly correlated with adverse outcomes in COVID-19 patients \cite{covidmodel}. Although lactic dehydrogenase was not part of our COVID-19 dataset, the temporal signatures of both C-reactive protein and lymphocytes have been accurately identified by \emph{Dynamic-KD} as predictors. This, in addition to age being marked as an important static feature, agrees with recent findings \cite{demographics}, showing that the signals jointly picked up by the framework's modules are coherent and well-aligned with clinical findings. 

\section{Discussion}
We developed and validated \emph{KD-OP}, an end-to-end pipeline for predicting adversity during hospitalisation. The pipeline comprises two stacked modules, each making predictions from a view of the patient's data: dynamic time-series and static features. The stacking of the pipeline's modules enables mimicking a clinician's approach to making prognostic decisions, by taking into account the interplay between the temporal signatures of a patient's physiology as well as time-invariant characteristics. By design, the pipeline is cognizant of the class imbalance natural to hospitalisation outcome data. It is trained and validated using stratified data that retains the original distribution of the real outcomes within the population. 

Model interpretability has been extensively linked with clinical utility and trust in clinical risk prediction systems \cite{interpretablemain}, where clinicians are generally reluctant to accept decisions guided by 'black-box' Machine Learning models \cite{interpretable1,interpretable2}. \emph{KD-OP} enables the post-hoc interpretation of its predictions by extending the idea of using attention weights to interpret neural network outcomes \cite{nips}. \emph{KD-OP}'s visualisation component accounts for the interplay between dynamic and static features in justifying the predictions made by the pipeline, which is a feature that derives directly from the stacked architecture. To our knowledge, this feature is not available in any existing hospitalisation outcome predictor.

We evaluated \emph{KD-OP}'s performance using real hospital data on three use cases representative of the diversity of electronic health records data. Using the pipeline to predict mortality and ICU admission/re-admission over 5-day, 7-day, 14-day, and 30-day intervals resulted in prediction accuracies exceeding 90\% in all mortality outcomes and most of the ICU admission/re-admission outcomes. 

\emph{KD-OP} is among the few models available in the literature that have been validated in ICU and non-ICU settings. Adversity prediction in ICU settings is a less challenging endeavour due to the high volume of frequently-recorded variables and the relative uniformity of patients with respect to the level of acuity. In a non-ICU setting, \emph{KD-OP} outperforms all existing models when considering the wide range of metrics needed to make an informed judgement about the models' predictive power. More importantly, \emph{KD-OP} is the only model that has been validated using PR-AUC, which measures the model's ability to predict minority (adverse) outcomes.

The high performance achieved by \emph{KD-OP} in predicting adverse outcomes in three diverse patient populations is a confirmation of the model's capacity in recognising \emph{context}. The model's ability to identify the most relevant features for a given diagnosis reflects what one would like to see in a general outcome prediction model, where the primary diagnosis serves as a guide to making a prognosis about a given patient. A disease-agnostic model such as \emph{KD-OP} could be built into the visual display of an EHR for all clinicians to use. The present challenge is that each hospital department has its outcome prediction scoring system, subsequently making it unrealistic to build over 30 distinct models into an EHR system. The generic nature of \emph{KD-OP}, coupled with high performance and visualisation capability, gives it a broader potential for integration in ICU and non-ICU settings. 

\emph{KD-OP} can be extended in a number of ways. First, it would be interesting to project the progression of the risk of adversity; we are currently developing a temporal risk score model to predict and visualise the risk of a given outcome on an individual level over time, using \emph{KD-OP} as the base model. Second, the current pipeline only supports classification outcomes. Existing targets include the prognosis of continuous outcomes such as worsening oxygenation or cardiac function. Also, the current framework strictly uses routinely collected clinical variables as predictors. Other types of data can be of high relevance to a given use case. For example, ECG signals are the predictors of choice for cardiology-related outcomes; X-ray images can positively improve predictive power in the case of COVID-19, etc. Although the stacked architecture has proven to be highly robust compared to parallel ensembles, it is intrinsically less flexible towards incorporating additional models, which renders extending the stacked model an interesting research problem. Finally, we are currently extending the framework's justification component to incorporate both the magnitude and direction of variable contribution to \emph{Static-OP} and \emph{Dynamic-KD}'s predictions. However, our ultimate goal is to extend beyond a correlation-based visualisation and into a decision-theoretical framework enabling the contextual selection of predicted risk based on the modular and overall framework performance.


\newpage
\bibliographystyle{plain}
\bibliography{ref}

\end{document}